%% file: main.tex
\documentclass{article}

\usepackage[nonatbib,final]{neurips_2021}

\usepackage{natbib}
\usepackage[utf8]{inputenc} 
\usepackage[T1]{fontenc}    
\usepackage{hyperref}       
\usepackage{url}            
\usepackage{booktabs}       
\usepackage{amsfonts}       
\usepackage{nicefrac}       
\usepackage{microtype}      

\usepackage{xcolor}
\usepackage{graphicx}
\usepackage{subcaption}
\usepackage{amsmath}
\usepackage{wrapfig}
\usepackage{bbm}
\usepackage[ruled,vlined]{algorithm2e}

\usepackage{todonotes}
\usepackage{multirow}
\usepackage{multicol}
\usepackage{array}
\input{math_commands.tex}

\setcitestyle{square,numbers}

\newcommand{\adversary}{generator}
\newcommand{\webnav}{learner}
\newcommand{\Adversary}{Generator}
\newcommand{\Webnav}{Learner}
\newcommand{\cminigrid}{cMiniGrid}
\newcommand{\cminiwob}{gMiniWob}

\newcommand{\WEDP}{PopRegret}
\newcommand{\WEDPb}{CoDE}

\title{Environment Generation for Zero-Shot Compositional Reinforcement Learning}

\author{%
  Izzeddin Gur, Natasha Jaques, Yingjie Miao, Jongwook Choi, \\ 
  \textbf{Manoj Tiwari, Honglak Lee, Aleksandra Faust} \\
  Google Research, Brain Team\\
  Mountain View, California, 94043 \\
  \texttt{\{\href{mailto:izzeddin@google.com}{izzeddin}, natashajaques, yingjiemiao, mjtiwari, sandrafaust\}@google.com} \\
  \texttt{\{jwook, honglak\}@umich.edu } \\
}

\begin{document}

\maketitle

\begin{abstract}
Many real-world problems are compositional -- solving them requires completing interdependent sub-tasks, either in series or in parallel, that can be represented as a dependency graph. Deep reinforcement learning (RL) agents often struggle to learn such complex tasks due to the long time horizons and sparse rewards. To address this problem, we present Compositional Design of Environments (CoDE), which trains a \Adversary\ agent to automatically build a series of compositional tasks tailored to the RL agent's current skill level. This automatic curriculum not only enables the agent to learn more complex tasks than it could have otherwise, but also selects tasks where the agent's performance is weak, enhancing its robustness and ability to generalize zero-shot to unseen tasks at test-time. We analyze why current environment generation techniques are insufficient for the problem of generating compositional tasks, and propose a new algorithm that addresses these issues. Our results assess learning and generalization across multiple compositional tasks, including the real-world problem of learning to navigate and interact with web pages. We learn to generate environments composed of multiple pages or rooms, and train RL agents capable of completing wide-range of complex tasks in those environments. We contribute two new benchmark frameworks for generating compositional tasks, compositional MiniGrid and gMiniWoB for web navigation. CoDE yields 4x higher success rate than the strongest baseline, and demonstrates strong performance of real websites learned on 3500 primitive tasks.
\end{abstract}

\vspace*{-0.1in}
\section{Introduction}
\vspace*{-0.1in}
Consider purchasing an airline ticket, logging in to a website, or buying movie tickets. These tasks can be completed by mastering a small set of basic manipulation skills (\textit{primitives}), such as entering an appropriate text in a fill-in field, or selecting a date, and combining them in different ways to form complex, \textit{compositional tasks} \cite{compositionalRLsurvey}. Humans can easily generalize between compositional tasks -- purchase a ticket on an airline or fill out a form that they have not seen before, even when the task carries over several pages -- but training autonomous agents to do this is far from straightforward. Yet, unlocking generalization across related tasks would pave the way towards autonomous agents that can automatically handle the details of completing wide variety of real-world user requests such as, ``Buy me a plane ticket to Los Angeles leaving on Friday''. The complexity and diversity of real environments make this a formidable challenge, especially due to the exponentially exploding action space and sparse rewards. 

Generalizing across compositional tasks is challenging for several reasons, including the \textit{tractability of training} \cite{compositionalRLsurvey}. Many compositional tasks require planning over an excessively long horizon while providing only sparse rewards, making it difficult for RL agents to learn.
Presenting easy tasks until the agent learns can be a solution, but it is not clear how to construct such a curriculum for complex compositional tasks \cite{compositionalRL}. Manually designing a pre-defined curriculum is tedious, and intractable. Domain randomization (DR) \cite{tobin2017domain,gur2018learning} does not tailor the difficulty of the task to the ability of the agent. Automatically generating tasks tailored to the agent's learning progress is a promising approach, but as we will show, current approaches (e.g. \citep{dennis2020emergent}) suffer from fundamental limitations in the context of generating compositional tasks. 

We present a new automatic curriculum generation algorithm, Compositional Design of Environments (CoDE), which jointly trains a \Adversary\ and a population of \Webnav\ agents. The \adversary\ is trained with a novel multi-objective reward function. Its first objective is to maximize the regret between the agents in the population and the best-performing agent, stabilizing regret estimation and making it less vulnerable to becoming stuck in local minima. Second, the \adversary\ is trained to 
adjust the task difficulty to match the \webnav s' proficiency using an explicit difficulty incentive designed for use with compositional tasks. In this way, the \adversary\ builds more challenging environments when the \webnav s are performing well, and reduces the difficulty when the \webnav s are struggling. We demonstrate that this difficulty incentive addresses degenerative edge cases present in prior work.

\begin{figure}[tb]
    \vspace{-0.3cm}
    \center
    \begin{tabular}{ccccc}
    \subfloat[Iter=100]{
        \includegraphics[width=0.15\linewidth]{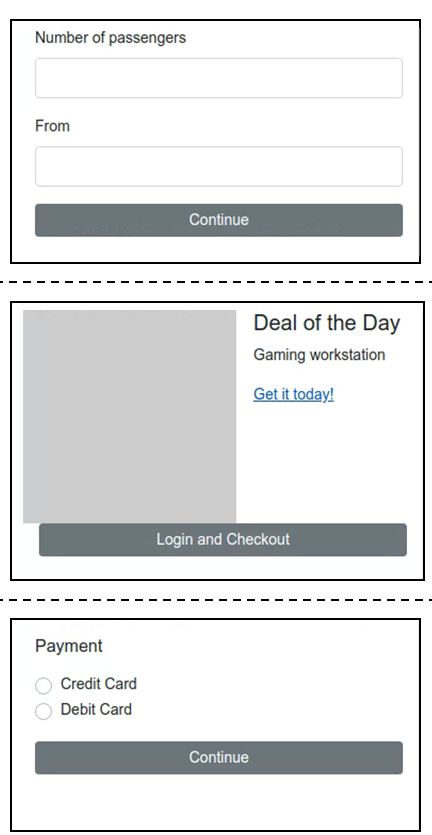}} &
    \subfloat[Iter=12000]{
        \includegraphics[width=0.17\linewidth]{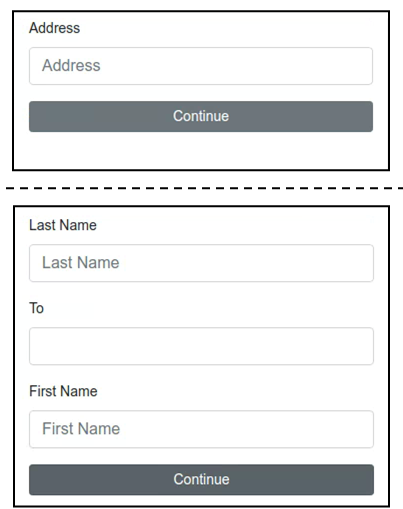}} &
    \subfloat[Iter=24000]{
        \includegraphics[width=0.16\linewidth]{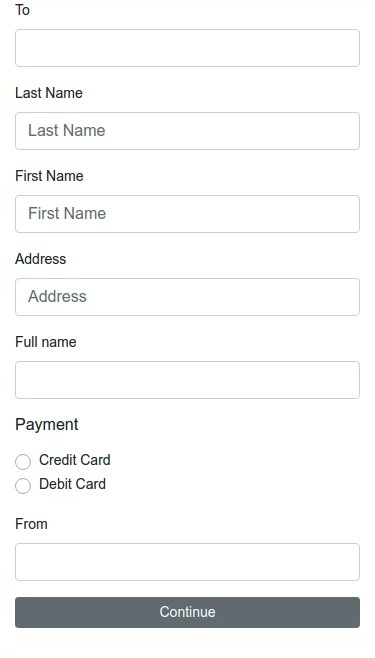}} &
    \subfloat[Web Nav. Test]{
        \includegraphics[width=0.17\linewidth]{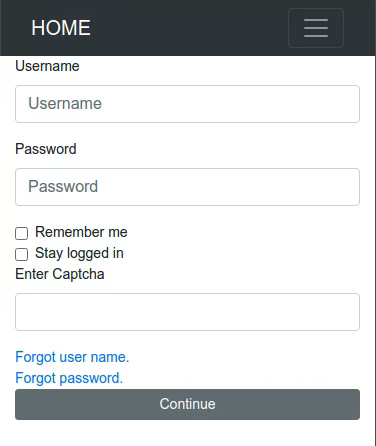}} &
     \subfloat[cMiniGrid Test]{
        \includegraphics[width=0.17\linewidth]{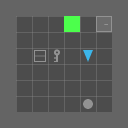}}
    \end{tabular}
    \caption{\small Environments examples. Generated Web pages become more complicated over the course of training (a-c), unseen test ``Login'' website (d) and "Task List" minigrid (e) environments. More in Appendix \ref{ap:example_designs}.
    }
    \label{fig:samplewebsites}
    \vspace{-0.5cm}
\end{figure}

By automatically searching for tasks on which the \webnav s are performing poorly, the \adversary\ makes the \webnav s more robust, enabling them to generalize to unseen tasks at test time. This ability is necessary for many real-world tasks, such as web navigation.
Prior work on web navigation relied on collecting demonstration data from a series of training websites \citep{shi2017world,liu2018reinforcement}, or training a separate policy for every single website \cite{gur2018learning}. These techniques cannot scale to a production-quality system, because they would require re-training the agents every time an airline company updated its site. In contrast, we show that \WEDPb\ produces robust agents with general form-filling and navigation skills that can be applied zero-shot to novel tasks at test time. 

Another challenge in learning compositional tasks is that the \textit{existing benchmarks do not match real-life complexities}. For example, real websites are orders of magnitude more complex than existing benchmarks \citep{shi2017world,liu2018reinforcement}. To address this problem, we introduce two new benchmark tasks for compositional task generation, which we are releasing in open-source. The first provides a way to automatically generate Minigrid \cite{gym_minigrid} navigation tasks which include subtasks such as locating a key to open a door. The second, generative MiniWoB (\textit{gMiniWob}), focuses on web navigation and manipulation (form-filling) tasks and enables a \adversary\ to construct increasingly complex form-filling websites, spanning multiple pages, out of common design primitives such as \textit{navigation bars}, \textit{product carousels}, \textit{item decks}, and \textit{item carts} (Figure \ref{fig:samplewebsites}). The evaluation environments in gMiniWoB are orders of magnitude more complex than MiniWoB, the prior web navigation benchmark \citep{shi2017world}.

This paper makes the following contributions. First, we formally introduce compositional tasks by drawing a connection to the Petri Nets graph formalism, and showing its relationship to POMDPs. We then analyze why prior techniques for automatically generating a curriculum of tasks are insufficient for compositional tasks, and propose a new algorithm, \WEDPb\, that addresses these weaknesses. We build two new automatic task generation environments spanning simple navigation tasks and web navigation, and release both in open-source. We demonstrate strong empirical results across both domains. In the context of web navigation, we show that \WEDPb\ generates a curriculum of increasingly challenging websites. Resulting agents successfully generalize to complex, unseen sites at test time, and without additional training solve wide range of form-filling tasks from flight purchases to login in. \WEDPb\ agents solve the most difficult tasks with $\approx 90\%$ success rate, 4x improvement over the strongest baseline. Lastly, we demonstrate that the method scales up to real websites.
The implementation of CoDE and gMiniWoB framework are available in open source at \url{https://github.com/google-research/google-research/tree/master/compositional_rl}.

\section{Related work}
\textbf{Compositional task representation:}
Compositional tasks \cite{agre1990plans} have been represented as a vector sum of primitive skills \cite{plan-arithmetic}, or sequence of subtasks \cite{oh2017zero}. The dependency graph representation \cite{sohn2018hierarchical,sohn2019meta} considers a task as a graph of subtasks equipped with dependencies as logical expressions. We relax the definition of the compositional tasks to represent real-world challenges. First, we allow primitives that do not lead to task completion. Second, we allow for partially observed environments. Third, we formalize the task distribution with a family of Petri Nets \cite{petrinet}, which have also been used to describe robot tasks \cite{petri-robot}, planning tasks \cite{petri-no-explode}, and workflows \cite{petri-workflow}. We use PNs with color and hierarchy, structured similarly to the workflows, but allow dangling nodes. \citet{mdp-pertinet} map PNs to MDPs, while we propose mapping between a set of PNs to a set of POMDPs.

\textbf{Curriculum Generation:} A variety of curriculum learning methods exist which do not generate whole environments. For example, \citet{florensa18a} train a Generative Adversarial Network (GAN) to generate a curriculum of goal images for a goal-conditioned policy to reach. Other methods focus on choosing which task out of a set of pre-defined tasks to present next \cite{narvekar2017autonomous,zhang2020automatic,feng2020novel}.
Another line of research studies creating curricula where a target task is already given \cite{leonetti2017,peng2018}.
Multi-agent training can also be effective for automatically generating a curriculum \cite{leibo2019autocurricula,matiisen2019teacher,graves2017automated,portelas2020teacher,sukhbaatar2017intrinsic,narvekar2018learning}. \citet{czarnecki18a} takes an alternative approach and generates a curriculum over agents instead of tasks. Finally, Active Domain Randomization \cite{mehta2020active,raparthy2020generating} learns to set the parameters of an environment simulator in order to provide a curriculum of increasingly difficult environments. In contrast, we focus on sequentially constructing an entire environment or task, composed of multiple subtasks, without a target task is explicitly given.

\textbf{Environment Generation:} 
POET \cite{wang2019paired,wang2020enhanced} generates terrains for a 2D walker with a population of environment-generating adversaries. 
\citet{campero2020learning} train a teacher to propose navigation tasks based on a length threshold.
Most closely related to our work is PAIRED \citep{dennis2020emergent}, which generates environments that induce maximal regret between a pair of agents. However, PAIRED only demonstrates results on simple gridworlds, and does not expand to complex, compositional tasks. We analyze several deficiencies in PAIRED that inhibit its use for the compositional setting, including a degenerate case where it fails to work at all. Our proposed algorithm addresses these deficiencies. To the best of our knowledge, we are the first work to attempt to automatically generate compositional tasks, and the first to apply environment generation to the problem of web navigation.

\textbf{Web navigation benchmarks and tasks:}
MiniWoB \citep{shi2017world} and MiniWoB++ \citep{liu2018reinforcement} environments contribute a fixed set of manually curated toy websites, and train agents using expert demonstrations for each website. This cannot scale effectively to cover the large variety of real-world websites, and cannot adapt to changing websites. Further, these methods failed to solve complex tasks such as flight booking or social media interaction. Using a manually scheduled curriculum \citep{gur2018learning} improves upon these results, but does not adapt to the agent's progress or generalize to unseen sites.
This work differs in several ways. First, we introduce a new framework, gMiniWoB, that allows generating complex websites on-the-fly with tunable difficulty levels. Additionally, we do not rely on any expert demonstrations to augment sparse rewards.
Most importantly, our web navigation agents generalize to unseen environments, an important requirement for real-world web navigation agents.

\vspace*{-0.1in}
\section{Compositional tasks problem definition}
\label{sec:definition}
\vspace*{-0.1in}
\begin{figure*}[tb]
\vspace{-1.cm}
    \center
    \subfloat[Primitives]{
        \includegraphics[width=0.24\linewidth]{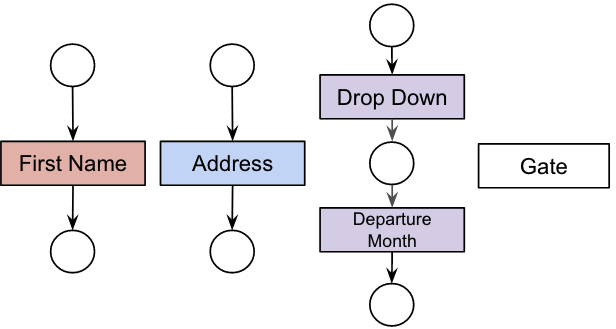}\label{fig:primitives}}
    \subfloat[Example 1]{
        \includegraphics[width=0.36\linewidth]{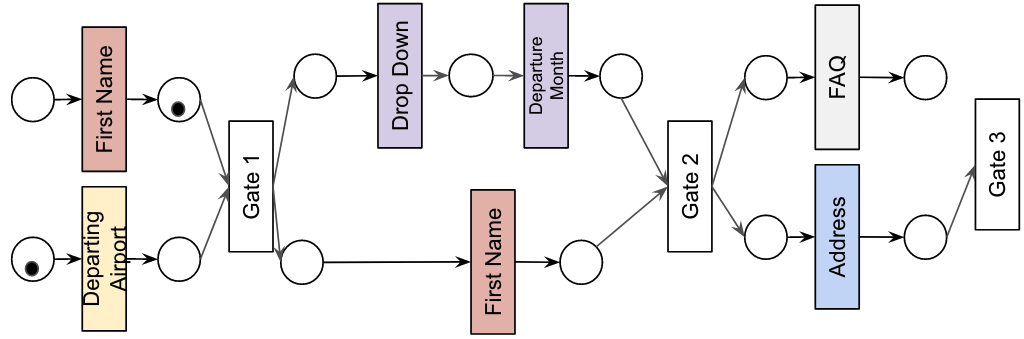}\label{fig:petri_ex1}}
    \subfloat[Example 2]{
        \includegraphics[width=0.36\linewidth]{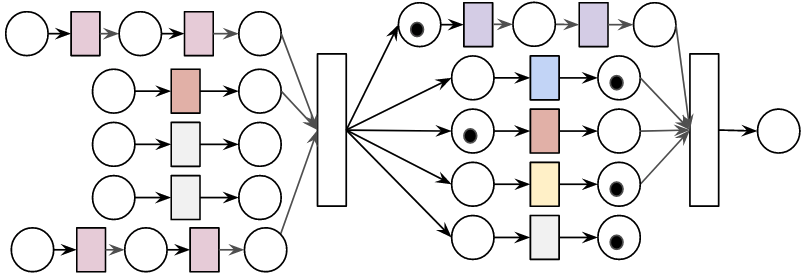}\label{fig:petri_ex2}}%
    \caption{\small Examples of compositional task primitives (a) and two different compositional tasks composed using the same primitives (b-c). 
    Note, passive primitives (in grey) that do not lead to the task progression and serve as a distraction for the agent. Difficulty budget for the \adversary\ is based on the number of primitives used. }
    \label{fig:petris}
    \vspace{-0.4cm}
\end{figure*}

Compositional tasks are composed of subtasks, which in turn are composed of primitive skills. This type of problem can be formalized using a task dependency graph, in which dependencies between the subtasks are represented as edges (see Figure \ref{fig:petris}). 
We model tasks in which the agent must complete several subtasks in order to trigger a phase transition in the environment, such as locating a key and unlocking a door in order to enter a new room, or completely navigating and correctly manipulating one web page in order to transition to the next one. 

We assume that the navigation agent uses standard POMDP setup -- a tuple $(\mathcal{S,A,T,O},r,\gamma)$, where $s \in \mathcal{S}$ are states, $o \sim \mathcal{O}(s)$ are partial observations, $a \in \mathcal{A}$ are actions, and $\gamma \in [0,1)$ is a discount factor. At each timestep $t$, the agent selects action $a_t$ according to its policy $\pi$, receives reward $r(a_t,s_t)$, and the environment transitions according to $\mathcal{T}(s_{t+1}|s_t,a_t)$. Agents seek to maximize their \textit{return}, the discounted sum of rewards: $R(\pi) = \sum_{t=0}^T \gamma^t r_t(a_t,s_t)$.

We use Petri Nets (PNs) \cite{colored-petrinets,hperti} framework to formally define \textit{learnable compositional tasks}.
PNs are directional graphs consisting \textit{place} and \textit{transition} nodes, and easily model different dependencies such as sequential, concurrent, choice, etc. \textit{Places} are system states; \textit{transitions} are points in the process that takes the system to the next state. \textit{Edges} determine the dependencies between nodes. 
At execution time, PNs have tokens that propagate through the net between places and determine the system state.
In PNs with colors (CPN) \cite{colored-petrinets} tokens are assigned a value, and Hierarchical PNs \cite{hperti} replace subnets with transitions.

We define a \textit{primitive} as a workflow (a special type of PN) with multiple places/transitions where there is no parallelism (see Figure \ref{fig:primitives}). They can be combined in parallel or serial using another transition (ex: Figure \ref{fig:petri_ex1}).
The set of colors, $C$, represents data semantics that agent needs to place in the environment to complete the task (i.e. flight departure or return dates).
In the form filling task, agent needs to complete all the relevant fields before proceeding to the submit button that presents the next page. To that end, we introduce a \textit{gate}, a special transition state that completes a phase and moves the system to the next (e.g. move between the pages in web navigation, or move between rooms in navigation tasks).
\textit{Learnable compositional tasks}, is a family of a directed acyclic PNs with colors induced with a set of primitives $P_C$.

\textit{Learnable tasks} described above map compositional tasks to POMDP $(\mathcal{S,A,T,O})$ that define RL navigation agents, and \adversary\ architecture which allows the \adversary\ to generate tasks only with the structure defined by PN.
See Appendix \ref{app:petri} for more details on PetriNets and their connection to POMDPs.

\vspace*{-0.1in}
\section{Environment generation with minimax regret analysis}
\label{sec:failure_modes}
\vspace*{-0.1in}
To train robust agents able to learn complex tasks, it would be helpful to generate a curriculum of tasks that are designed to be just outside of their current skill level.
This requires training a \Adversary\ policy $\pi^G$ to construct an environment $\mathcal{E}$ that will challenge a \webnav\ policy $\pi^L$ (e.g. \cite{wang2019paired}). To find tasks that are solvable, but where the \webnav\ is weak, we can train the \adversary\ to maximize the \textit{regret}. Regret is the difference between the return obtained by policy $\pi^L$ and the return that would have been obtained with the optimal policy $\pi^*$ in the same environment $\mathcal{E}$: $\textsc{Regret} = R^\mathcal{E}(\pi^*) - R^\mathcal{E}(\pi^L)$. Because the optimal policy is not known \textit{a priori}, PAIRED \cite{dennis2020emergent} approximates the regret by introducing a third \textit{antagonist} agent, with policy $\pi^A$, and computing the regret with respect to it: $\mathcal{J}^{\textsc{PAIRED}} = R^\mathcal{E}(\pi^A) - R^\mathcal{E}(\pi^L)$.
In order to generate a viable curriculum, regret minimization relies on the assumption that the reward function includes an incentive to complete the task more efficiently. In this case, the regret will be highest for easy tasks which could be completed in a few steps by the optimal policy, but which the \webnav\ policy $\pi^L$ fails to complete. Thus, regret is a useful objective for inducing a curriculum when it is possible for the \adversary\ to build impossible environments, and the reward includes incentives to complete the task in fewer timesteps.

\begin{wrapfigure}{r}{0.4\textwidth}
\small
\centering
  \includegraphics[width=\linewidth]{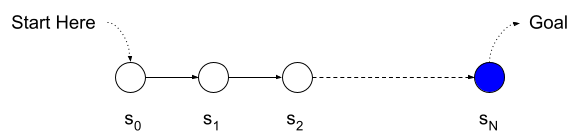}
\caption{\small In chain environments, it becomes exponentially more difficult to reach the goal via random exploration as $N$ increases.}
\label{fig:chain}
\end{wrapfigure}
\textbf{Degenerate case.} What happens if the assumptions listed above are violated? For example, consider the case where there is only a binary reward $r \in \{+1, -1\}$ depending on whether the agent correctly performs a subtask or not, and the \adversary\ is constrained such that it cannot construct impossible environments. These assumptions are both true in the case of the compositional task graphs we study in this paper. Consider the simplest possible task graph, the Chain MDP pictured in Figure \ref{fig:chain}. 
The agent only receives a positive reward if it reaches state $s_N$, and from state $s_i$ it can only reach state $s_{i-1}$ or $s_{i+1}$.
The probability of reaching $s_N$ through random exploration is $P_{reach}(N) = O(p^N)$ (see Appendix Section \ref{app:reachprob} for the proof), which decreases exponentially as more states are added.
By our assumption, we know it is possible to reach $s_N$ and so the reward for the optimal policy is $R(\pi^*)=1$. Then, maximizing $\textsc{Regret} = 1 - R^\mathcal{E}(\pi^L)$ reduces to minimizing the probability that the \webnav\ obtains a reward, i.e. $P_{reach}(N)$. This can be accomplished by continuing to increase $N$. For compositional tasks, this means the \adversary\ will add as many elements to the environment as possible. In this case, maximizing regret will not lead to a curriculum, and could actually slow learning more than randomly generating environments with Domain Randomization. 

\textbf{Sparse rewards.} Even if the assumption that $R^{\mathcal{E}}(\pi^*)$ is higher for easier environments is met, the regret signal used to train the \adversary\ can still be extremely sparse. Consider the case where both the \webnav\ and antagonist agent cannot solve the task. In this case, they never receive a positive reward, their rewards are equal, and the regret is zero. This gives the \adversary\ no signal with which to learn to decrease the difficulty of the environment to enable them to start learning. Once again, the \adversary\ is no better at generating a curriculum than randomly sampling environments. 

\textbf{Lack of convergence and stalled training.} The proof that the PAIRED algorithm will produce a \webnav\ agent with minimal regret relies on the assumption that the game will reach a Nash equilibrium \cite{dennis2020emergent}. However, in practice gradient-based multi-agent RL has no convergence guarantees, is highly non-stationary, and will often fail to converge \citep{mazumdar2019policy,mazumdar2019finding}. If the game does not converge, then PAIRED minimizes regret with respect to the antagonist's policy, which only forces the \webnav\ to be as good as the antagonist. If the antagonist fails to improve, or reaches a local optimum, then the curriculum will stall and the \webnav\ cannot continue to improve. 

\vspace*{-0.1in}
\section{Compositional Design of Environments (\WEDPb)}
\label{sec:code_algo}
\vspace*{-0.1in}
We present a new algorithm, Compositional Design of Environments (\WEDPb), for automatically generating a curriculum of compositional tasks. \WEDPb\ trains a \adversary\ to construct compositional environments out of primitives, and then uses those environments to train a population $\mathcal{P}$ of \webnav\ agents. The rewards obtained by the \webnav s across environments are used to compute a multi-objective reward function to train the \adversary\ to produce an effective curriculum.

\textbf{Population-based regret estimation:}
Despite issues with its estimation in prior work, regret is a useful incentive for environment generation because it encourages the \adversary\ to explore towards environments that promote variation in performance, indicating they have learning potential. Thus, the first component of our algorithm is a new, more flexible population-based estimate of regret (PopRegret) that is not subject to the problem of stalled training if the game converges to local optima. For each environment $\mathcal{E}$ constructed by the \adversary, each agent $p \in \mathcal{P}$ collects $M$ episodes in the environment. The regret is the difference between the average score of the population, and the score of the policy that achieved the highest average return over the $M$ episodes. Let $R^{\mathcal{E}}_m(\pi^p)$ be the return obtained by policy $p$ in episode $m$. Then:
\begin{align}
    \mathcal{J}^{\textsc{PopRegret}} &= \max_p \E_m[R^{\mathcal{E}}_m(\pi^p)] - \frac{1}{K} \sum_{i=1}^{|\mathcal{P}|} \E_m[R^{\mathcal{E}}_m(\pi^i)] \label{eq:flex_regret}
\end{align}
As long as any agent has higher performance than any other agent, the objective will continue to identify environments in which there is learning potential, preventing training from stalling. As the agents continue learning, the best-performing agent in the population more closely approximates the optimal policy, providing a stronger estimate of the regret. Further, the regret estimate is smoother and more consistently positive than the regret obtained via the maximum of a pre-selected agent, providing a more stable signal for the \adversary\ to learn to optimize.

\begin{figure}[tb]

    \centering
    \includegraphics[width=0.75\linewidth]{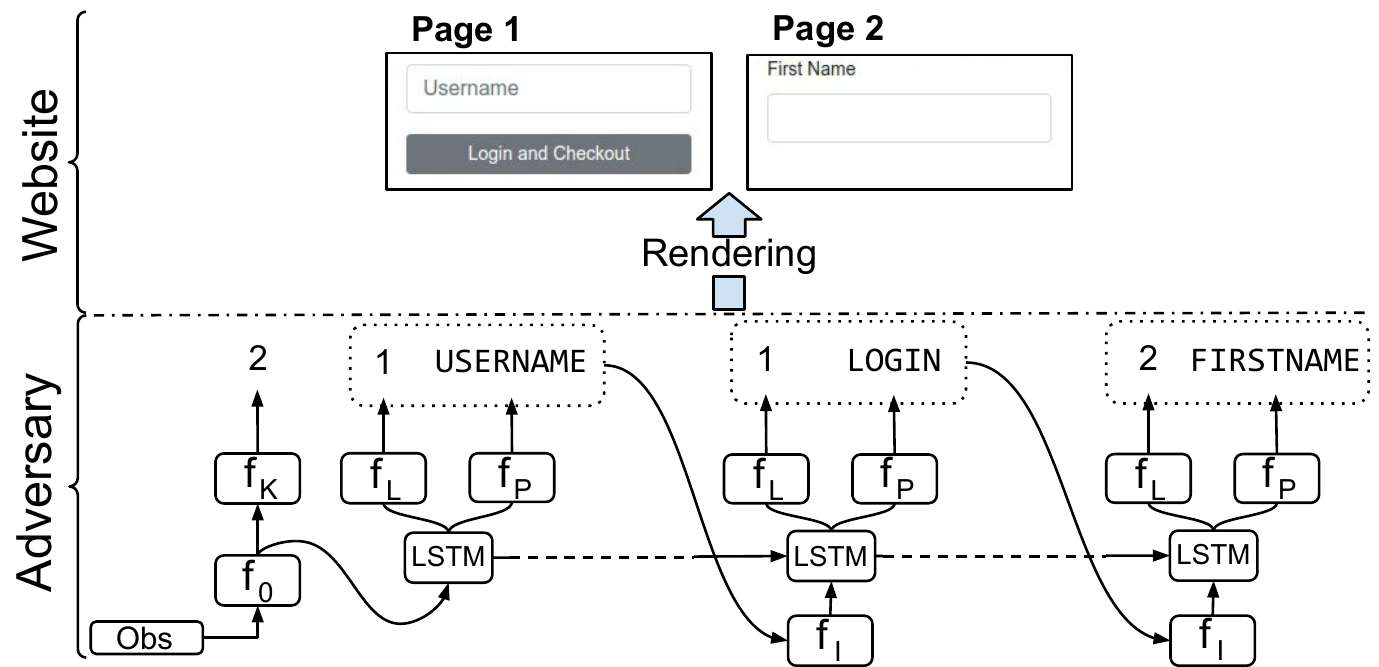}
    \caption{\small A sample rollout of the \adversary\ for a compositional web navigation task. An initial observation (Obs) is given at the beginning of the rollout. $f_0$, $f_K$, $f_P$, and $f_I$ networks encode initial observation, creating subtasks, primitives, and encoding LSTM inputs. In practice, we also use an additional independent network $f_L$ to create page indices where each primitive is assigned.}
    \label{fig:rollout}
    \vspace{-0.3cm}
\end{figure}

\textbf{Difficulty budget:}
Even with a more stable estimate of the regret, training can still be slow and unstable if the rewards are sparse, and potentially subject to the degenerate case identified in Section \ref{sec:failure_modes}.
To address these issues, we add an incentive for the \adversary\ to tune the difficulty of the environment to the current skill level of the agents.
Let $N$ be the number of primitives added to the task graph, and $p = \argmax_{p\in\mathcal{P}} R^\mathcal{E}(\pi^p)$ be the agent with the highest score in environment $\mathcal{E}$. Then:
\begin{align}
    \mathcal{J}^{\textsc{Difficulty}}(\mathcal{E}) &=  (\mathbbm{1}[R^\mathcal{E}(\pi^p) > \beta] - \mathbbm{1}[R^{\mathcal{E}}(\pi^p) < \delta])N/N_{max} \label{eq:budget}
\end{align}
where $\mathbbm{1}$ is the indicator function and $\delta < \beta$ are reward thresholds for failure and success, respectively.
Thus, if the best-performing agent gets a high score on the task, the \adversary\ gets a reward based on the number of primitives it added to the environment,  $\frac{N}{N_{max}}$. However, if all the agents are performing poorly, the \adversary\ is penalized according to the number of primitives it added and receives $-\frac{N}{N_{max}}$. This makes adding more primitives risky, unless the \adversary\ can be sure that at least one agent in the population can solve the environment. It is thus incentivized to create tasks that challenge all but the best-performing agents in the population, while still decreasing the difficulty when the agents are not learning.

\textbf{\Adversary\ model:}
\label{sec:adversary}
The \adversary\ is a decoder designed to construct an environment $\mathcal{E}$ by placing a set of primitives $a \in \mathcal{A}$ within a set of subtasks $[\mathcal{W}_1, \cdots, \mathcal{W}_{K}]$. 
We assume a fixed maximum number of subtasks $K$, and allow for a up to $N$ primitives to be added to the environment. To control for complexity, empty subtasks with no primitives are allowed. The \adversary's policy $\pi^G(\mathcal{E} | o^G)$ is conditioned on an initial observation $o^G \sim N(0,I)$ sampled from the standard normal distribution, which allows it to diversify its design distribution (similar to a GAN generator \cite{goodfellow2014generative}). It then constructs the environment according to:
\begin{equation}
    \pi^G(\mathcal{E} | o^G) = \pi_{\mathcal{E}_w}(k | K)\prod_{i=0}^n{\pi(a_i|a_{0:i-1}, j)}
\end{equation}
where $a_i$ corresponds to a primitive. The \adversary\ first encodes $o^G$ with a feed forward network $h_0=f_0(o^A)$. The encoding $h_0$ is passed to another network $f_K$ that outputs a Categorical distribution $Cat(0, K)$, from which the number of subtasks is sampled $k\sim Cat(0, K)$. Given $k$ and initial state $h_0$, the \adversary\ executes an autoregressive LSTM policy to sample primitives $a_i$, through a network $f_P$ (see Appendix \ref{ap:adversary_network_details} for more details). 
Thus, the generator is a stochastic, open-loop policy.
Note that there is a \texttt{SKIP} action that does not add a new primitive, enabling the generator to reduce complexity. The finished design $\mathcal{E}$ is sent to a renderer module to generate the environment. Figure \ref{fig:rollout} shows an example of how the \adversary\ constructs an environment which the renderer converts to a website. Note that the \adversary\ is domain independent, and creates a generic compositional task or PN; indeed, we use the same model for both experiment domains. 

\textbf{Difficulty budget implementation:} Rather than directly counting the number of primitives for the difficulty, we replace $N$ in Eq. \ref{eq:budget} with the amount of probability the \adversary\ places on the \texttt{SKIP} action throughout generating the environment: $\hat{N} := -\sum_{i=1}^n{\log  \pi_{\mathcal{E}}(a_i=\texttt{SKIP}|a_{0:i-1})}$. 
This allows for a more direct control over \adversary\ policy to optimize difficulty. We also scale Eq. \ref{eq:budget} by the reward received by the best-performing agent to provide further signal to the generator.

\textbf{Algorithm:}
Algorithm \ref{alg:bflexible} shows the \WEDPb\ training procedure.
Both the \adversary\ and \webnav\ agents are trained using RL, specifically A2C \citep{a2c} with entropy regularization.
For every training step, the \adversary\ constructs an environment $\mathcal{E}$, and the agents $p \in \mathcal{P}$ collect $M$ trajectories within $\mathcal{E}$.
The \webnav\ agents are trained using the standard task related reward. To train the \adversary, we use the following multi-objective loss function that encourages the adversary to control the complexity of the environment by presenting ``just-the-right challenge" for the agents in $\mathcal{P}$, where $\alpha$ is a hyperparameter:
\begin{align}
    \mathcal{J}(\mathcal{E},\mathcal{P}) = (1-\alpha) * \mathcal{J}^{\textsc{PopRegret}}(\mathcal{E},\mathcal{P}) + \alpha * \mathcal{J}^{\textsc{Difficulty}}(\mathcal{E},\mathcal{P})
    \label{eq:loss} 
\end{align}

\SetKwComment{Comment}{$\triangleright$}
\SetCommentSty{itshape\color{black!60}}
\begin{algorithm}[tb]
 \textbf{Initialize} policies for \adversary\ $\pi^G$, and agents $\pi^p, \forall p \in \mathcal{P}$ \;
 \For {all training iterations}{
    $\mathcal{E}_w \sim \pi_G(\mathcal{E}|o^G), o^G \sim N(0,I)$  \Comment*{\Adversary\ constructs environment}
    \For {$p=1,\cdots,|\mathcal{P}|$}{ 
        \For {$m=1,\cdots,M$}{
             $R^{\mathcal{E}}_m(\pi^p)$ $\longleftarrow$ Collect rewards for agent $p$ in environment $\mathcal{E}_w$\;
         }
        RL update for agent $p$ using collected experience \Comment*[r]{Train \webnav}
     }
    Compute $\mathcal{J}^{\textsc{Regret}}$ using Eq. \ref{eq:flex_regret}\;
    Compute $\mathcal{J}^{\textsc{Difficulty}}$ using Eq. \ref{eq:budget} \;
    RL update for \adversary\ using Eq. \ref{eq:loss} as the reward \Comment*{Train \adversary}
 }
 \caption{\small \WEDPb: Joint training of the \adversary\ and \webnav\ agents.}
 \label{alg:bflexible}
\end{algorithm}

\vspace*{-0.1in}
\section{Benchmark environments}
\vspace*{-0.1in}
We contribute two open-source frameworks for compositional tasks construction out of a set of primitives: \textit{Compositional MiniGrid (\cminigrid)} and \textit{Generative MiniWoB (\cminiwob)}.

\textbf{\cminigrid} is a compositional extension of MiniGrid navigation environments \cite{gym_minigrid}, in which agents interact with various objects. \cminigrid\ contains the following subtasks: \textit{pick up the key}, \textit{unlock the door}, \textit{pick up the ball}, \textit{open the box}, \textit{drop the ball in the box}, \textit{reach the goal}. The \adversary\ chooses subtasks to add, and the renderer places objects. The problems have sparse rewards which are only given for completing a subtask, which each require many steps. 
\textbf{\cminiwob} is a generative extension of miniWob and miniWob++ web navigation benchmarks, extending them to multi-page navigation and richer websites.
\cminiwob\ enables generation of web navigation tasks consisting of multiple web pages (subtasks), each composed of primitives (see Figure \ref{fig:petris}). The renderer handles linking the pages via output primitives (e.g. by adding a submit button).
Each primitive is represented using a Document Object Model (DOM) tree, and rendered using HTML. 
gMiniWoB implements 40 common web primitives, such as navigation bars, carousels, item decks, web forms, item carts, dropdowns, etc. 
The order in which primitives are added defines how a webpage will be rendered for the agent, facilitating controllable generation with a rich design set.
Every primitive changes the DOM structure when the agent interacts with it.
$26$ of the $40$ primitives are active, while the rest are passive (i.e. they do not relate to the task, like an ad banner). 
When the \adversary\ adds an active primitive, the renderer adds the corresponding piece of data to the agent's instruction set, increasing the complexity of the task.
For example, adding a `First Name' text box also adds a new \textit{``firstname''} field into the instruction.
See Appendix \ref{ap:design_primitives} for all the primitives, and Appendix \ref{ap:test} for the test set. Note that with 40 possible primitives across 10 web pages, \cminiwob\ provides the ability to generate $\approx 10^{14}$ distinct web navigation tasks.

As in \cite{gur2018learning}, the web-navigating RL agent receives a text instruction containing various fields of information, and must interact with DOM elements, fill in the necessary information, and navigate through the pages to complete the task. The agent observes the DOM tree of the current page. Its actions are a tuple (element, field) that denotes inputting the field into the element. For example, in a flight booking task, given the instruction \{\texttt{"Departure Date": "Friday", Destination Airport: "Los Angeles (LAX)", "Address": ...}\}, the agent must first pick the textbox labeled "dest", find the corresponding field in the instruction (Destination Airport) and type the value ("Los Angeles (LAX)").
The agent receives a small penalty each timestep to encourage efficient navigation. Correctly filling out information within the page results in a positive reward, normalized over the total number of fields in the instruction (e.g. if there are $F$ fields, the agent receives reward of $1/F$). Agents receive a reward of 1.0 for completing the task, and -1.0 otherwise.

\section{Evaluations}
\textbf{Implementation:} Following \citet{gur2018learning}, the \webnav\ policy is an LSTM based DOM tree encoder and a feed forward network to encode instruction fields.
The policy outputs a joint distribution over elements and fields by measuring pairwise similarities between element encodings and instruction fields.
We compute the state-value by using the marginal distribution of elements as attention weights over element encodings and passing the context vector through a feed-forward network. \WEDPb\ is implemented using ACME \cite{acme} with TensorFlow \cite{tf} open-source libraries. The training is done on a single CPU, requiring about a week of training. The results reported are averaged over 5 seeds.

\textbf{Baselines:} We compare to: a) \textit{PAIRED} \cite{dennis2020emergent}: for a fair comparison, we limit the size of the \WEDPb\ population $|\mathcal{P}|=2$, since PAIRED uses two agents to estimate regret; b) \textit{ALP} \cite{portelas2020teacher}: we use absolute difference between rewards of a single navigator at timestep $t$ and $t-1$ as the reward for the adversary without any difficulty bugdet; c) \textit{Domain Randomization (DR)}: \cite{sadeghi2016cad2rl,tobin2017domain} we sample the number of subtasks $k$ and the primitives from uniform distributions.
d) \textit{Curriculum Learning (CL)}:
based on the state-of-the-art web navigation RL agent \citet{gur2018learning}. We adapt this method to zero-shot environment generation where we are not given a specific website but a set of design primitives.
We randomly sample each primitive w.r.t. a probability $p$ where $p$ is initialized with a small number and scheduled to reach $1.0$ during training; d) Learning from Demonstrations for web navigation from DOM inputs (LfD DOM) \cite{shi2017world} and image (LfD Visual) \cite{liu2018reinforcement}.

\textbf{Web navigation evaluation:}
We evaluate our models on MiniWoB \citep{shi2017world}, and a suite of test environments implemented in gMiniWoB (`Login', `Enter Address', `Flight Booking', `Enter Payment', and `Shopping' websites).
Each environment comes with 4 different difficulty levels by gradually adding more primitives.
These environments are never explicitly presented to agents during training, so performance measures how well agents can generalize to unseen websites at test time. 
To train \WEDPb\ we use 40 hand-engineered primitives, included with \cminiwob\, which are also used as building blocks for the test environments.
Finally, we conduct an experiment based on real websites, where the \adversary\ chooses from among 3500 primitives scraped from the web. The primitives were extracted from 200 different websites in the password change, food ordering, and shopping domains. We evaluate agents' performance on a held out set of real websites, never presented during training.
\subsection{Results}
\label{sec:results}
\begin{figure*}[tb]
\center
\small
    \subfloat[Legend]{
      \includegraphics[width=0.10\linewidth]{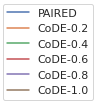}
    }%
    \subfloat[Complexity]{
      \includegraphics[width=0.28\linewidth]{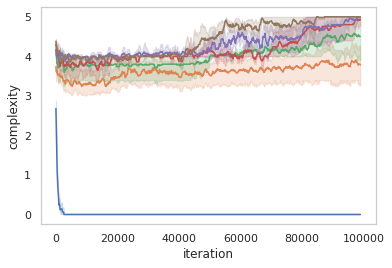}
      \label{fig:minigrid_complexity}
    }%
    \subfloat[Regret]{
      \includegraphics[width=0.28\linewidth]{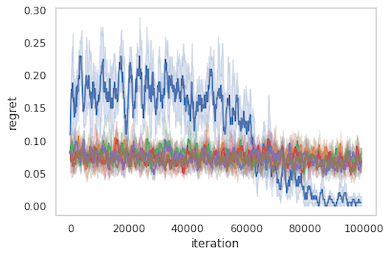}
    }%
    \subfloat[Reward]{
      \includegraphics[width=0.28\linewidth]{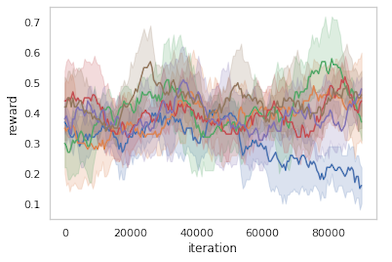}
    }%
    \caption{\small \cminigrid\ experiments. Due to sparse rewards, the PAIRED algorithm fails to generate complex environments or train agents with high reward. Adding difficulty incentive (with increasing $\alpha$) enables both.}
    \label{fig:minigrid_results}
\end{figure*}

\begin{figure*}[t]
\center
\small
    \subfloat[Difficulty level 1]{
      \includegraphics[width=0.24\linewidth]{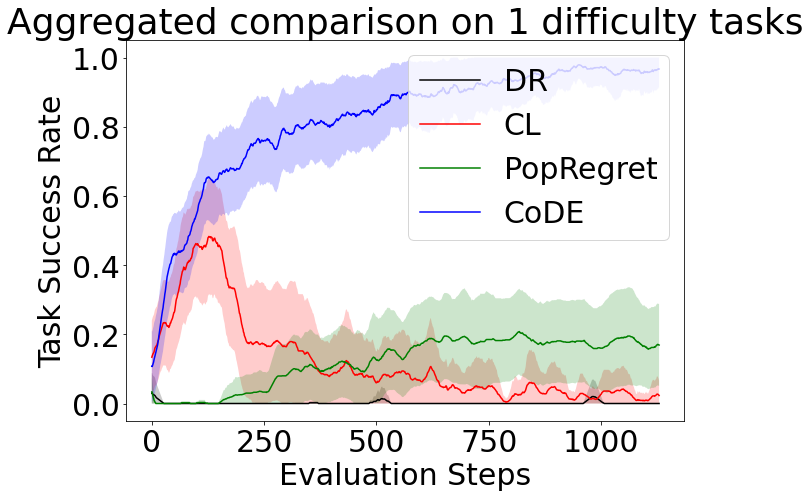}
    }%
    \subfloat[Difficulty level 2]{
      \includegraphics[width=0.24\linewidth]{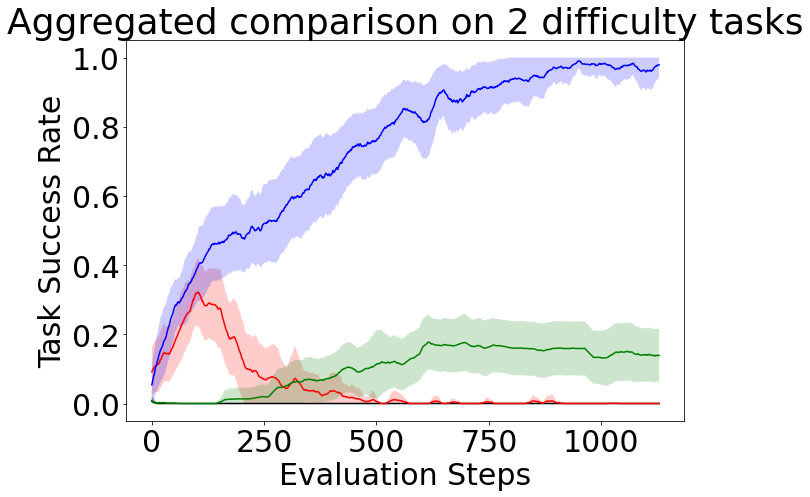}
    }%
    \subfloat[Difficulty level 3]{
      \includegraphics[width=0.24\linewidth]{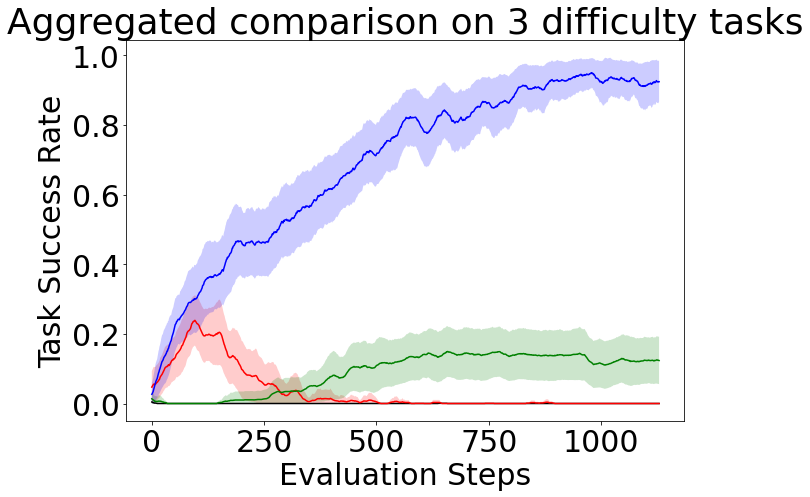}
    }%
    \subfloat[Difficulty level 4]{
      \includegraphics[width=0.24\linewidth]{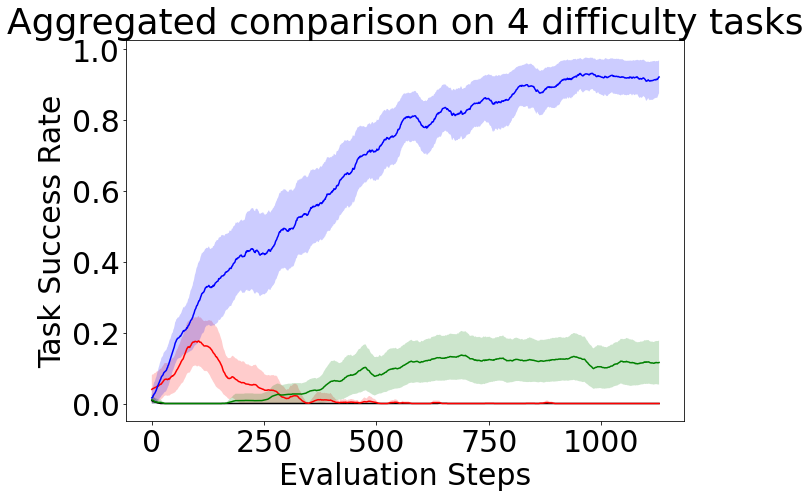}
    }%
    \caption{\small Aggregated task success rate comparison of \WEDPb\ and baseline models on test environments with increasing difficulty levels. See Appendix \ref{ap:detailed-results} for detailed results.}
    \label{fig:tasksuccessaggregated}
\end{figure*}

\begin{table}
\begin{center}
\small
\begin{tabular}{ c|c|c|c|r|r|r|r|r|r}
 \textbf{Env.}& \textbf{Task} & \textbf{DOM} &  \textbf{Instr.}  & \multicolumn{2}{c|}{\textbf{LfD}} & \multicolumn{1}{c|}{\textbf{CL}} &
 \multicolumn{1}{c|}{\textbf{ALP}}&
 \multicolumn{1}{c|}{\textbf{PAIRED}}& \multicolumn{1}{c}{\textbf{\WEDPb}} \\
 \textbf{}  & & \textbf{Size} &  \textbf{Size}  & DOM\cite{liu2018reinforcement} & Vis.\cite{shi2017world} & \multicolumn{1}{c|}{\cite{gur2018learning}} &
 \multicolumn{1}{c|}{\cite{portelas2020teacher}} &
 \multicolumn{1}{c|}{\cite{dennis2020emergent}} & \\
 \hline
 &password  & 11 & 1& 0\% & 100\% & 100\% && 100\% \\ 
 MiniWoB & enter-text & 6 & 1  & 0\% & 100\% & N/A  &\multicolumn{1}{c|}{N/A}&\multicolumn{1}{c|}{N/A}& 100\%  \\ 
 \cite{shi2017world}& dynamic  & 6 & 1   &0\% & 100\% & 100\%  &   & 100\%\\ \hline
&login & 35 & 5   & & & 3\%& 23\%& 20\% & \textbf{92\%}\\ 
    &address & 38 & 7  &  &  & 0\%& 15\% & 8\% &\textbf{98\%}   \\ 
 gMiniWoB& payment & 49 & 5 &\multicolumn{1}{c|}{N/A}&\multicolumn{1}{c|}{N/A} & 0\% & 7\%& 16\%& \textbf{93\%} \\ 
    &flight &60&7 &  & & 0\%& 2\% & 20\%&\textbf{95\%}  \\ 
    &shopping  &183 & 12 && & 0\%& 6\% & 6\% & \textbf{95\%} \\ \hline 
\end{tabular}
\end{center}
 \caption{\WEDPb\ performance on MiniWoB form-filling and \cminiwob\ test environments compared to several baselines. DOM Size and Instruction Size illustrate the complexity of the environment. Learning from Demonstration (LfD) and Curriculum Learning (CL) methods train a separate model for each test environments, and the results are reported from the previous publications. PAIRED, ALP and \WEDPb\ train a single model that is evaluated across all benchmark tasks. \WEDPb\ generalizes to all tasks, and outperforms all baselines.}
 
 \label{table:miniwob}
\end{table}
\textbf{Overall performance:} \WEDPb\ outperforms the baselines across the benchmarks, both in web navigation and \cminigrid. \WEDPb\ is at least 4x more successful at completing web navigation tasks than the strongest baseline (see Table \ref{table:miniwob}), and reaches more than 90\% task success across all difficulty levels (Figure \ref{fig:tasksuccessaggregated}).
On \cminigrid\ environments, \WEDPb\ has nearly 3x the success rate of PAIRED, solving 45\% vs 16\% of tasks (Figure \ref{fig:minigrid_results}), demonstrating \WEDPb's versatility across very different domains.
We present additional results and ablation study in Appendix \ref{ap:ablation}.

\textbf{Regret minimization with sparse reward:} 
In \cminigrid\ rewards are sparse, and frequently no agents in the population receive reward, resulting in zero regret and absence of the training signal. As predicted, under these conditions maximizing the regret leads to poor performance (Figure \ref{fig:minigrid_results}). 
Training stalls, and the agents do not achieve high reward. In contrast, adding the difficulty objective significantly improves results. 

\begin{figure*}[tb]
\small
\centering
    \subfloat[Payment]{
      \includegraphics[width=0.24\linewidth]{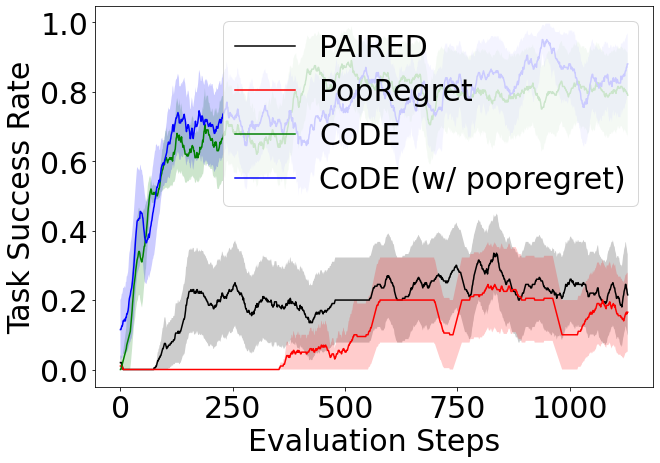}
    }%
    \subfloat[Shopping]{
      \includegraphics[width=0.24\linewidth]{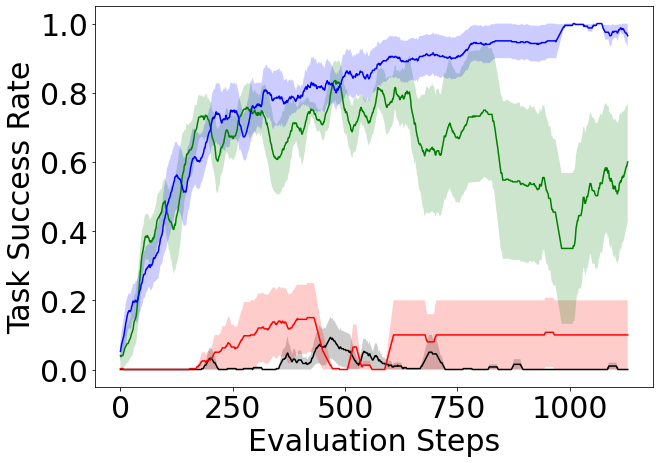}
    }%
    \subfloat[Flight booking]{
      \includegraphics[width=0.24\linewidth]{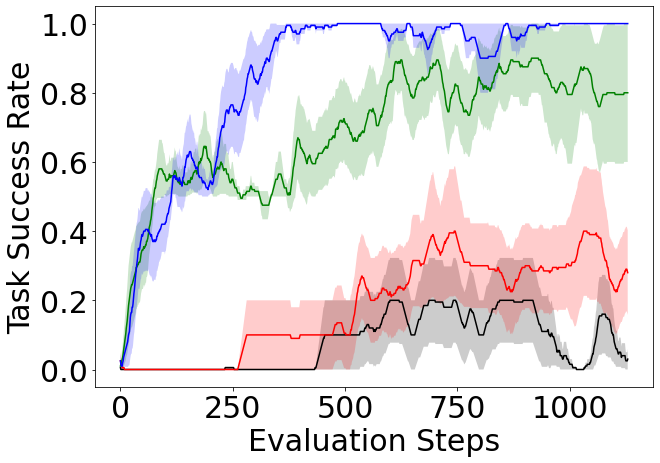}
    }%
    \subfloat[Primitives]{\includegraphics[width=0.24\linewidth]{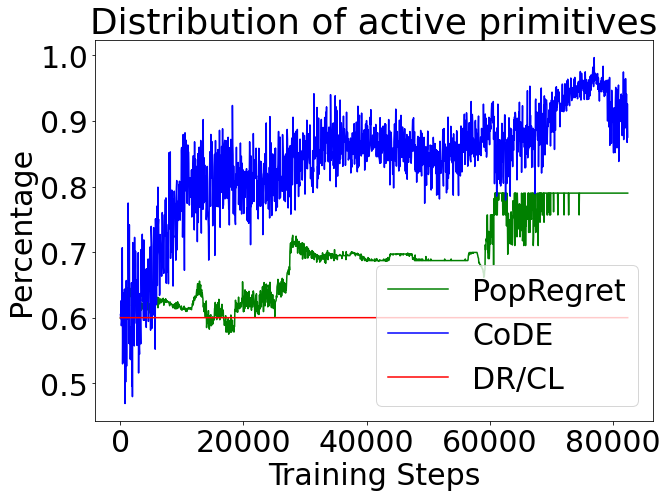}\label{fig:primitivedistribution}}
    \caption{\small Comparison of PAIRED \citep{dennis2020emergent} and \WEDPb\ averaged over 4 difficulty levels. (f): Percentage of active primitives over training steps (see Appendix \ref{ap:primitive_frequencies} for more details).}
    \label{fig:combinedpbt}
\end{figure*}

\textbf{Importance of difficulty budget}:
Difficulty incentive leads to a large improvement in performance of \WEDPb\ over the baselines on the web navigation tasks (Figures \ref{fig:combinedpbt} and \ref{fig:tasksuccessaggregated}). When agents are struggling to learn and have very similar rewards, the regret becomes very small. This uninformative signal makes it difficult for the regret-based methods (PAIRED and PopRegret) to learn. On the other hand, \WEDPb\ provides a clear signal to the \adversary\ that improves performance significantly over all tasks under consideration. Detailed results are in Appendix \ref{ap:detailed-results}, and further ablation studies in Appendix \ref{ap:weight_comp} investigate the effect of the budget weight $\alpha$.


\textbf{Importance of better regret estimation}: 
While the difficulty incentive provides a large boost to performance, it has a limitation in that it may not always be straightforward to calculate the difficulty of a generated task (for example, if the number of primitives added does not correspond directly to difficulty). In this case, it is necessary to rely on more general objectives such as regret, and stablizing the regret objective becomes important. 
We demonstrate, in two ways, that the proposed population-based regret estimate in Eq.\ref{eq:flex_regret} (\WEDP) provides an advantage over the regret estimation in PAIRED (Figure \ref{fig:combinedpbt}). First, while PAIRED training stalls, especially for difficult tasks such as shopping and flight booking, \WEDP\ agents learn more effectively. Second, we assess the contribution of \WEDP\ to the \WEDPb\ algorithm by comparing \WEDPb\ with no PopRegret (i.e. using only the difficulty incentive) to the full algorithm. While relying solely on the difficulty incentive is sufficient for easier tasks like payment, it does not perform as well as \WEDPb\ in more complicated tasks. 
\textbf{Curriculum generation:} \WEDPb\ produces a curriculum of emerging complexity, estimated by the number of active and passive primitives. \WEDPb\ starts around 60\% random selection of primitives, and gradually generates more primitives while improving agent's  performance (Figure \ref{fig:primitivedistribution}). Even as the complexity of the environments continues to increase, \WEDPb\ agents still perform consistently well without degrading performance (Figure \ref{fig:tasksuccessaggregated}).
CL's performance drops significantly due to ignoring agents' skill level and making environments that are too challenging for agents to complete, therefore ignoring their ability to learn. 
DR performs poorly because the randomly generated environments are so complex that the agents never even begin learning.

Next, both number of active and passive primitives increases (Appendix \ref{ap:primitive_frequencies}), and the distribution of the primitives shifts to more complex and relevant primitives later on (Appendix \ref{ap:primitives}). 
Learning a web page with more passive primitives is a relatively easier task than a page with more active primitives, because passive primitives either add noise and should be ignored by the agents, or are used by agents only to navigate to another page.

Active primitives not only increase DOM tree size, but the number of instruction fields as well, making the matching between elements and instruction more challenging.
\WEDPb\ agents continue to improve performance thanks to the ability to tune the difficulty of the environments to agents' proficiency. 

\begin{wraptable}{r}{0.4\textwidth}
\small
\begin{tabular}{c | c |  c| c|c} 
  & 1 & 2 & 3 & 4 \\
 \hline
 Login & 62\% &  63\% & 55\% & 44\% \\
 \hline
 Address & 36\% & 25\% & 29\% & 38\% \\
 \hline
 Payment & 23\% & 18\% & 12\% & 9\% \\
 \hline
 Shopping & 44\% & 36\% & 34\% & 31\% \\
 \hline
 Flight & 29\% & 25\% & 31\% &23\% \\
 \hline
\end{tabular}
\caption{\small CoDE performance on real websites across tasks and difficulty levels.}
\label{table:real}
\end{wraptable}
\textbf{Real website evaluation:} \WEDPb\ successfully scales to real-world web navigation with  \adversary\ choosing from a almost 100 times larger set of primitives, yielding $\approx 10^{31}$ possible environments.
The trained agents generalize to unseen environments (Table \ref{table:real}) constructed out of real web data.
As the distribution of real primitives mainly focus on username, password, and address fields, we observe a high evaluation success on the Login and Address tasks.
Even though we used the same primitives for the Flight as above results, we observe a significant drop in performance which can be attributed to skewness of the primitive distribution.

\section{Broader impact, limitations, and future work}
PN formalism (Appendix \ref{app:petri}) is useful beyond this work, as it exposes compositional task topology. We use expanded workflow model, which although expressive for a rich set of tasks seeded with only a handful of primitives, is still only one topological class of tasks. Further investigation is needed into the correlation between compositional task toplogies and methods. Do these methods work for tasks with alternate routes? Our preliminary investigation shows that the difficulty budget is not appropriate for tasks without a goal state (terminal state in PNs), regardless if the goal is not-reachable, or the task is open-ended. What methods can enable learning compositional tasks with difficult topologies? 

Next, we evaluated the model trained with real website primitives on test set with hand-engineered primitives, with surprising results: success rate ranging from 90\% on address tasks to 0\% in a simple MiniWob enter-text. While expecting generalization across primitives is a tall ask, the results show that in some cases it is attainable. What are the circumstances when the primitives generalization is possible, and how can we develop models that expand the repertoire of primitives?

\vspace*{-0.1in}
\section{Conclusion}
\vspace*{-0.1in}
We present, Compositional Design of Environments (CoDE), a novel algorithm for zero-shot generalization compositional task learning. First, the Petri Nets formalism enables definition of the learnable task built from the few selected primitives, algorithmic creation of RL agents and \adversary\, and provides tools for toplogocial analysis of the learnable tasks. Second, we introduced two new objectives for curriculum generation: an improved, population-based estimate of regret, and a difficulty objective tailored to the problem of generating compositional tasks. Next, a proposed domain-agnostic \adversary\ architecture builds new hierarchical tasks out of primitives. To facilitate research on compositional tasks and environment generation, we controbuted two open-source benchmarks.
Finally, we demonstrated that \WEDPb\ generates a curriculum of emerging complexity, successfully trains agents outperforming baselines 4x, and results in a \textit{single} RL agent capable of filling out generic forms and completing variety of tasks on real websites.

\ifdefined\isaccepted
\section*{Acknowledgements}
We thank Sungryull Sohn for providing help on Playground environments and discussions.
\fi

\bibliographystyle{plainnat}
\bibliography{main}

\clearpage
\appendix

\newpage
\newpage
\section{Appendix}

\subsection{Petri Nets formalism}
\label{app:petri}
Petri Nets (PNs), directional graphs consisting \textit{place} and \textit{transition} nodes, model concurrent systems with synchronous and asynchronous processes. \textit{Places} are system states; \textit{transitions} are points in the process where a user or an agent interacts with the system to perform an action and transition the system to the next state. \textit{Edges} determine the dependencies between nodes. 
At execution time, PNs have tokens that propagate through the net between places. When a token reaches a place adjacent to a transition, if the transition is enabled and fires, the token is consumed, and new one is placed at the successor place. Token placement determines the system state.
PNs easily model different dependencies: sequential, concurrent, choice, fork, join, mutual exclusion etc.
In PNs with colors (CPN) \cite{colored-petrinets} tokens are assigned a value, and Hierarchical PNs \cite{hperti} replace subnets with transitions. 

We focus on tasks that are compositions of subtasks, sets of primitives that need to be completed in order to proceed to the next stage. 
A \textit{primitive} is a (simple or serial) workflow-net with colors and multiple places/transitions where there is no parallelism (see Figure \ref{fig:primitives}). They can be combined in parallel or serial using another transition.
The set of colors, $C$, represents data semantics that agent needs to place in the environment to complete the task (i.e. flight departure or return dates). 
We assume that for each color in $C$ there is at least one primitive associated with it. Let $P_C$ be set of all primitives wrt color set $C$. For example, primitives are widgets that correspond to different ways of inputting departure date (as a fill-in field, or a selection from a calendar). Primitives contain an initial place (a state before the agent manipulates the environment), manipulation sequence, and end place (a state of the environment post interaction). In the instance of departure field primitive, an initial place is an empty date field, and the end place is date field with a departure field from the profile (Figure \ref{fig:primitives}).  
In the form filling task, agent needs to complete all the relevant fields before proceeding to the submit button that presents the next page. To that end, we introduce a \textit{gate}, a special transition state that completes a phase and moves the system to the next (e.g. move between the pages in web navigation, or move between rooms in navigation tasks). We also partition the primitives in \textit{active}, which make progress towards task completion, and \textit{inactive} that are distractions and do not contribute towards the progress.
\textit{Learnable compositional tasks}, used here, is a family of a directed acyclic hyper PNs with colors induced with a set of primitives $P_C$ and the following properties.
PN has at least one gate. All primitives sampled from $P_C$ must be reachable from a gate, but only active primitives must be predecessors to a gate. 
A \textit{page} is a subnet consisting of all primitives leading out of a gate. 

\textit{Learnable tasks} described above map compositional tasks to POMDP $(O,A,T,R)$ that define RL navigation agents, and \adversary\ architecture which allows the \adversary\ to generate tasks only with the structure defined by PN.  

POMDP hidden state is a PN configuration computed as location of all tokens. The POMDP transition function maps directly to place $\rightarrow$ transition $\rightarrow$ next place in PN.
Observations consist of the set of colors, which provide the data for the task and are renderings of all primitives in a current page. Each transition node in PN task maps to a POMDP action. Available actions directly match to enabled transitions. Each action is represented as a tuple (element, field) that denotes acting on the element using the field as an input; i.e. typing the value of the field into the element.
Finally, the rewards associated with the transitions as follows. At every time step a transition fires. Agent receives a small penalty each timestep to encourage efficient navigation. Emission of new tokens after a successful firing results in a positive potential-based reward for the agent, normalized over the total number of transitions. Agents receive a task success reward (1.0 or -1.0) when PN reaches the final state or times out.
As an example, consider a flight booking task where the agent is given an instruction \{\texttt{"Departure Date": "Friday", Destination Airport: "Los Angeles (LAX)"}\}.
The agent first picks a field (e.g. destination airport) and finds the corresponding text box in the page; then the corresponding value (``Los Angeles (LAX)") is typed into the text box.
If this value is correct, the agent receives a positive reward of $1 /  2 $ where $2$ is the number of fields in the instruction. \Webnav\ RL agent is parameterized by a network $\pi(a_t | s_t; \Theta_i)$, that maps an observation $s_t$ to output action distribution to maximize the cumulative discounted reward, i.e.,  $O=\sum_{t=0}^{T}{\gamma^t r_t}$ where $r_t$ is the reward at time step $t$, $\gamma$ is a discount factor, and $T$ is the length of an episode.

To create \adversary\ the networks $f_P$ and $f_L$ build the task by sampling next primitive $a_i$ directly from the PN set of primitives, and its page locations $b_i$. Given the structure of the compositional task that we used here, location and identity of each primitive is determined with its id, page location, and time step. Compositional tasks with different topologies will need more sub-networks to determine the location of the selected primitive.  

PNs define the structure of learnable tasks induced with primitives, and map directly to POMDPs. The PN formalism allows us to reason about and sample related tasks in a principled way. POMDPs define a learning problem for the RL agent, which acts on the PNs to complete the task. Our goal is to train a single RL agent that solves the set of POMDPs. In the rest of the paper, \adversary\ designs PN compositional task and the corresponding POMDP, and the \webnav\ agent learns to solve it. 

\subsection{Probability of successfully reaching the goal in a Chain MDP}
\label{app:reachprob}

Consider the following chain MDP (Figure \ref{fig:chain}) where the agent starts at the leftmost state ($s_0$), can move right or left, the goal ($g=s_N$) is the rightmost state, and the reward is \{+1, -1\} depending on if the goal is reached within $N+2L$ steps.
Let's assume that initially, $p$ is the probability of taking a right action.
Reaching the goal at state $N$ via random exploration is
\begin{align}
    P_{reach}(N) &= \sum_{t=0}^{L}{P(\text{N+t right action and t left action})} \\
                &= \sum_{t=0}^{L}{C(N+2t, t)p^{N+t}(1-p)^t} \\
                &=p^N \sum_{t=0}^{L}{C(N+2t, t)(p(1-p))^t} \\
                &\leq p^N(1+p-p^2)^L 
\end{align}
where $L\geq N$ and the last line comes from $(1+x)^n=\sum{C(n,k)x^k}$, $C(n+2k, k) \geq C(n, k)$ for every $n>0$. In the simplest case where $L = 0$, this becomes $p^N$.

\subsection{Additional Experiments}
\label{ap:ablation}
In Table \ref{table:ablation}, we present ablation studies for various hyper-parameters in CoDE.
We experiment with positive and negative reward thresholds $\beta=\delta$ and different numbers of rollouts $M$ per training iteration.
For each ablation, we also sample $\alpha$ from $\{0.8, 0.9\}$.

$\beta=\delta$ is an important hyper-parameter that needs careful tuning for the best performance but CoDE still outperforms baselines for different values. 
We observe that using a relatively high reward threshold (0.2) causes the adversary to become more conservative and increase complexity only when navigators are performing very strongly.
Using a larger number of episodes can decrease performance. 

\begin{table}
\centering
\begin{tabular}{c | c |  c| c|c | c} 
  & Login & Address & Payment & Shopping & Flight \\
 \hline
 CoDE & 92\% & 98\% & 93\% & 95\% & 95\% \\
 \hline
 CoDE ($\beta=-0.2,\alpha=0.8$) & 86\% & 100\% & 90\% & 96\% & 87\% \\
 \hline
 CoDE ($\beta=0.2,\alpha=0.8$) & 57\% & 24\% & 22\% & 33\% & 10\% \\
 \hline
 CoDE ($\beta=-0.2,\alpha=0.9$) & 82\% & 94\% & 82\% & 88\% & 70\% \\
 \hline
 CoDE ($\beta=0.2,\alpha=0.9$) & 50\% & 28\% & 43\% & 25\% & 15\% \\
 \hline
 CoDE ($M=4,\alpha=0.8$) & 66\% & 71\% & 69\% & 71\% & 49\% \\
 \hline
 CoDE ($M=4,\alpha=0.9$) & 84\% & 96\% & 94\% & 91\% & 85\% \\
 \hline
 CoDE ($M=6,\alpha=0.8$) & 62\% & 47\% & 65\% & 39\% & 27\% \\
 \hline
 CoDE ($M=6,\alpha=0.9$) & 60\% & 68\% & 54\% & 69\% & 34\% \\
 \hline
\end{tabular}
\caption{Ablation study of CoDE using different $\beta=\delta$ and $\alpha$ hyper-parameters. Using positive $\beta=\delta$ gives more conservative designs, CoDE is robust to different $\alpha$ values, and using larger number of episodes $M$ gives worse performance.  }
\label{table:ablation}
\end{table}

\subsection{Training flow}
\label{ap:flow}
\begin{figure*}[tb]
\centering
     \includegraphics[width=\linewidth]{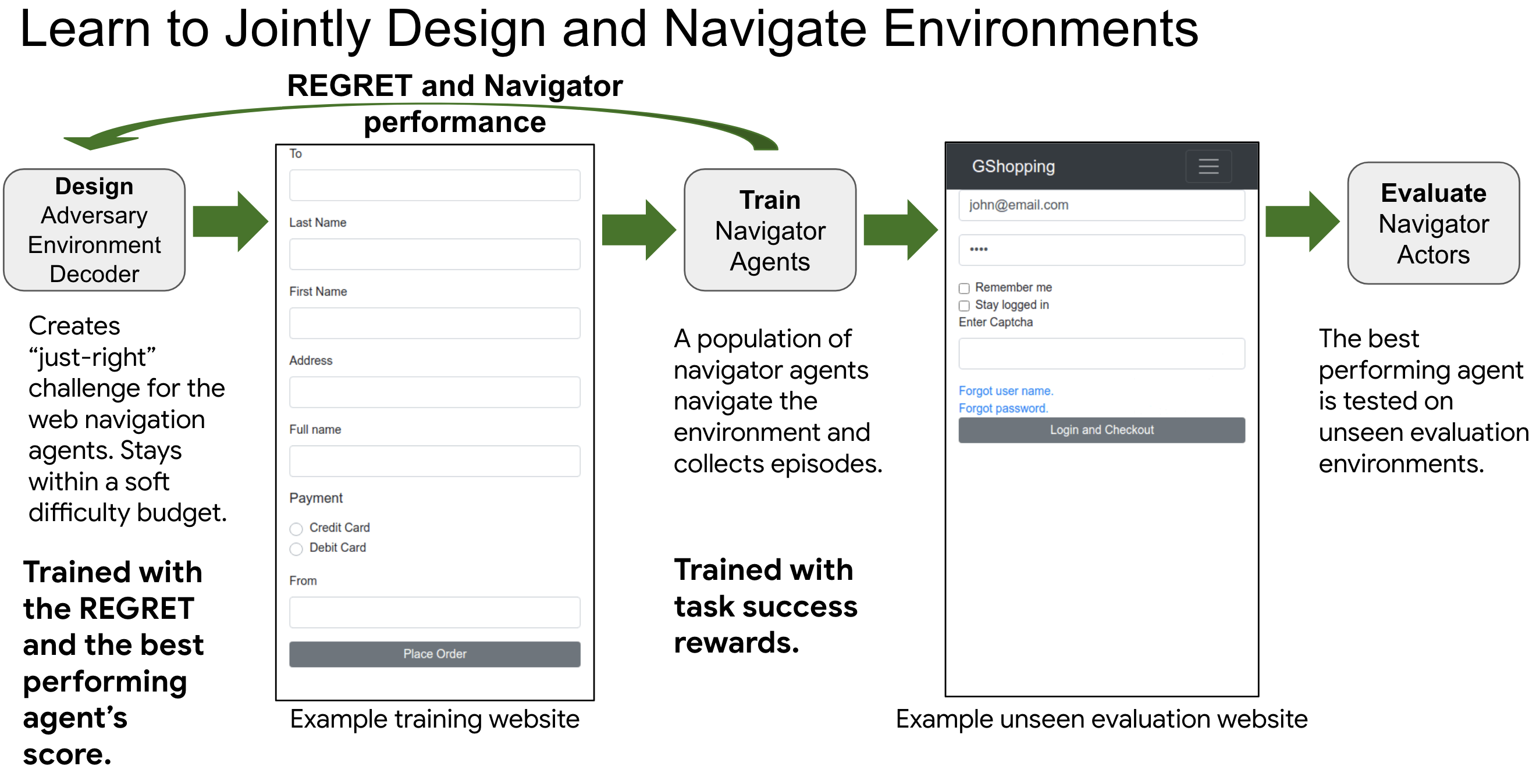}
    \caption{Training workflow. The adversary unrolls an environment, adding one element at the time of each page. That environment is passed on to a population of navigation agents under training. The navigators collect several episodes in the environment and their returns. The weight of the navigator agents are updated w.r.t. their returns. And the adversary weights are updated w.r.t. regret estimate and budget estimate using the best performing agent.}
    \label{fig:webnav_workflow}
\end{figure*}

In Figure \ref{fig:webnav_workflow}, we illustrate the high level workflow of the \WEDPb.

\subsection{Distribution of Primitives During Training}
\label{ap:primitives}
\begin{figure*}
    \subfloat[Early]{
      \includegraphics[width=1.0\linewidth]{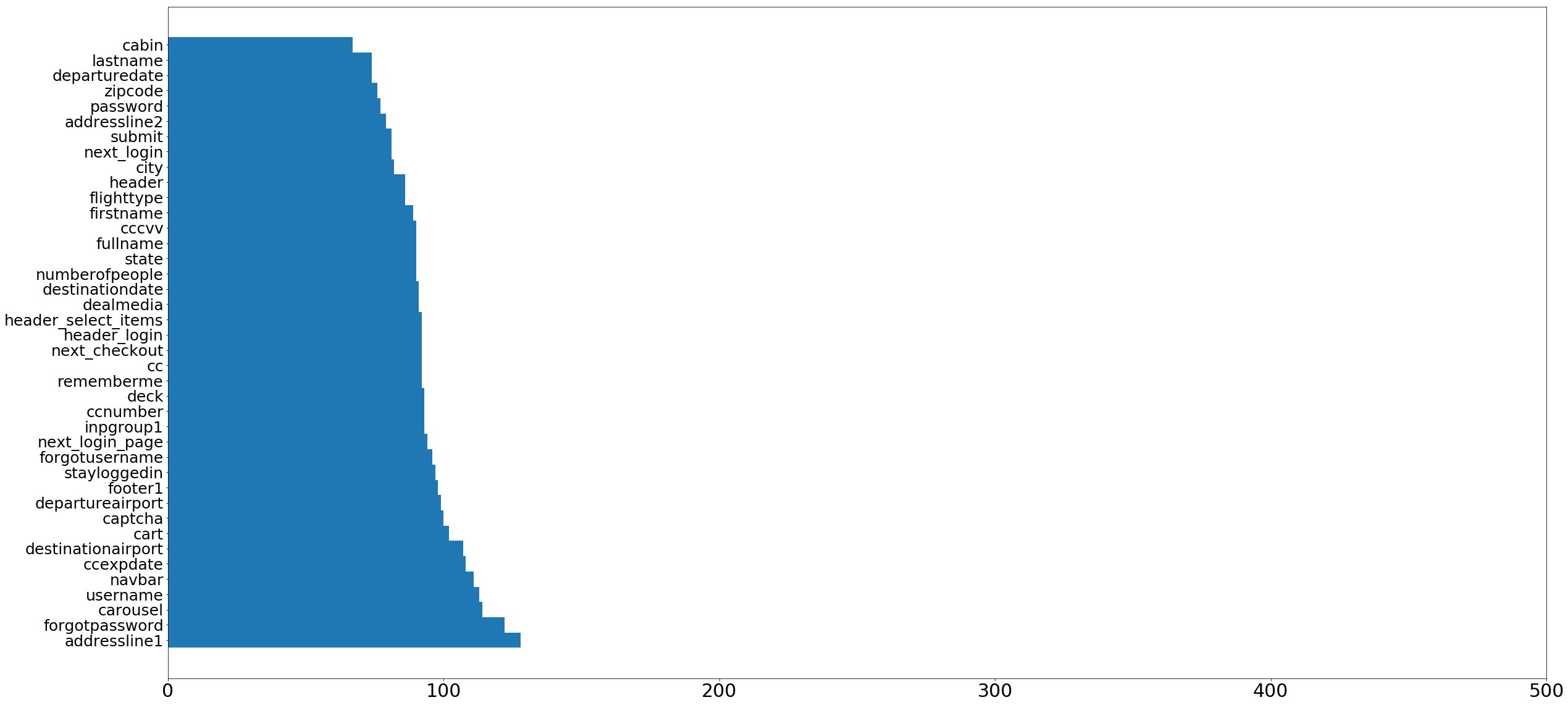}
    }\qquad
    \subfloat[Middle]{
      \includegraphics[width=1.0\linewidth]{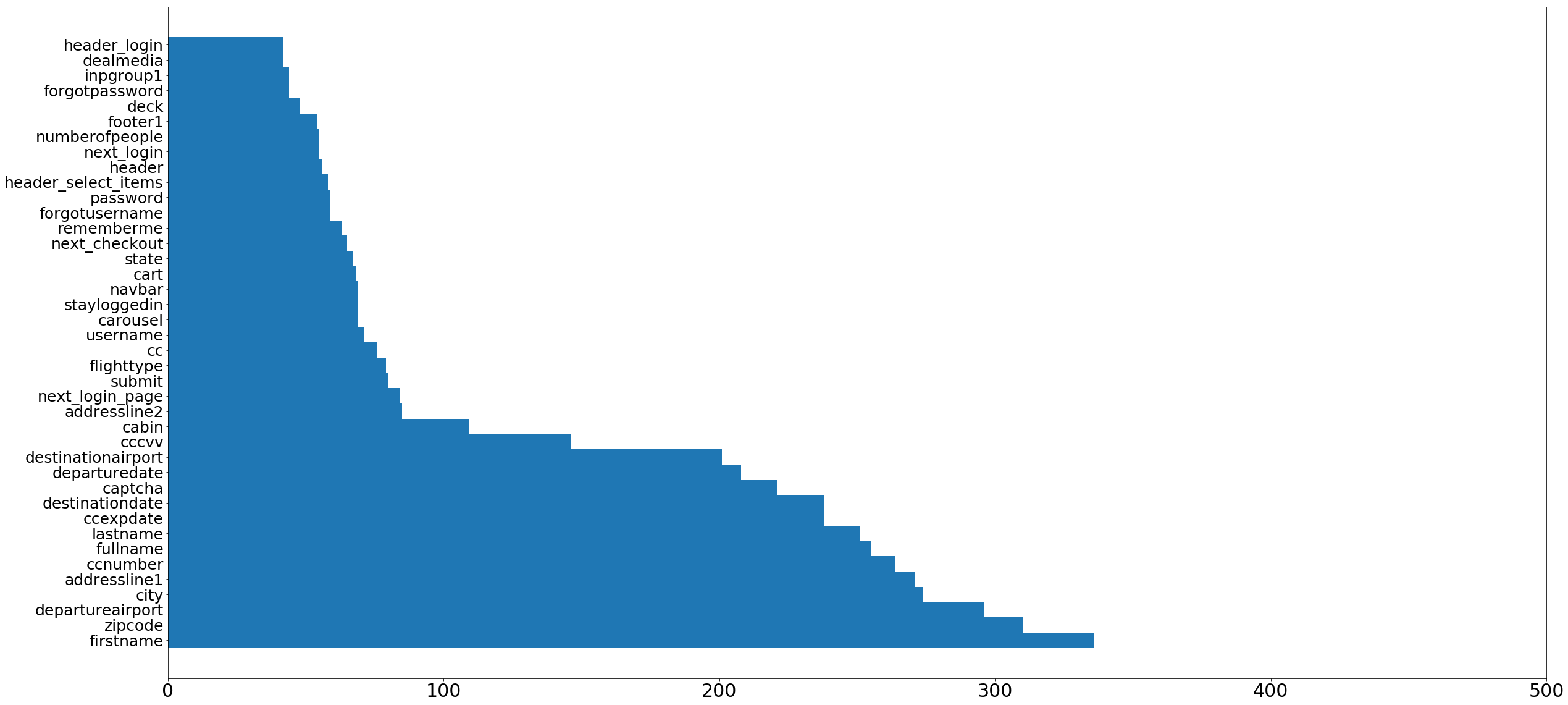}
    }\qquad
    \subfloat[Late]{
      \includegraphics[width=1.0\linewidth]{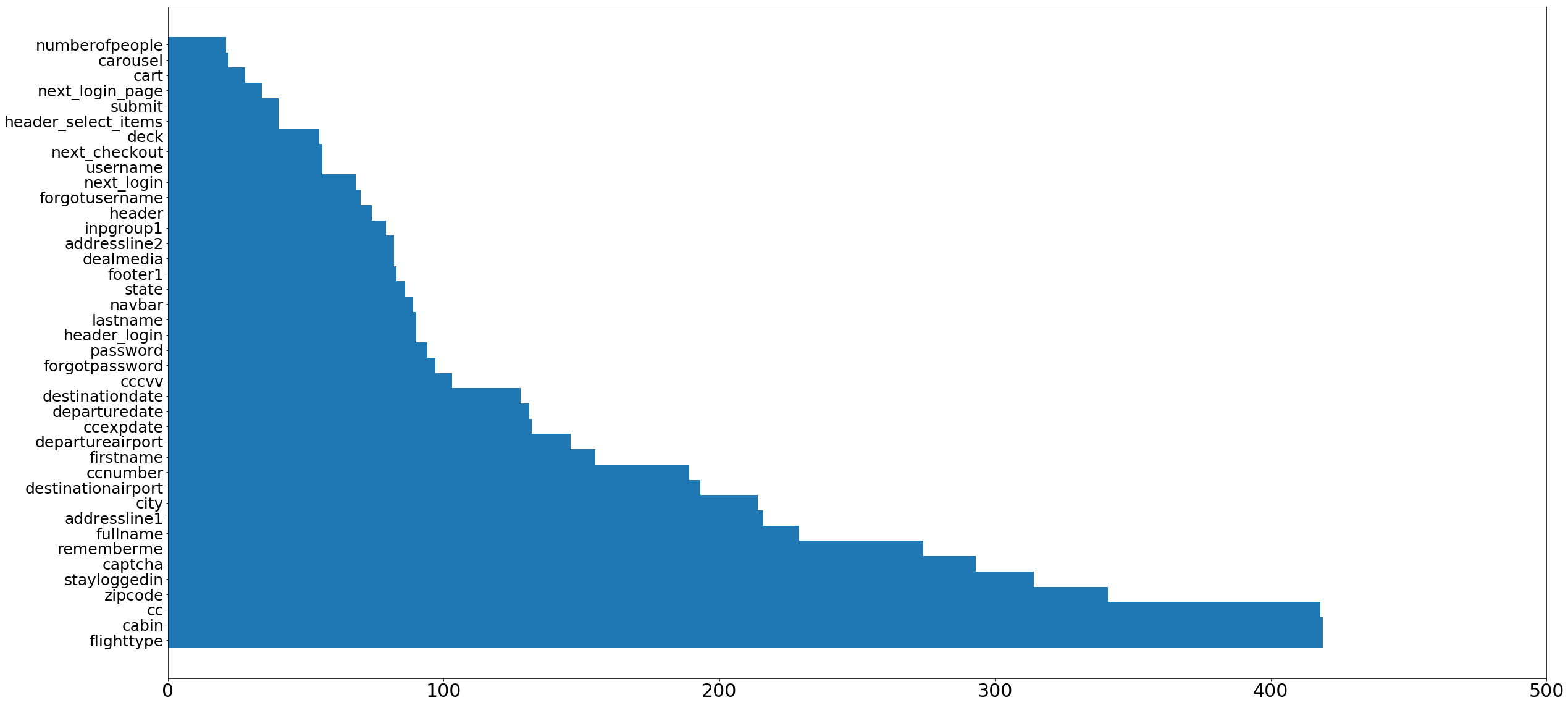}
      \label{fig:primitivehist3}
    }
    \caption{Histograms of primitives from early, middle, and late snapshots of the training.}
    \label{fig:primitivehist}
\end{figure*}
During training, the distribution of primitives become more skewed towards active primitives early on (as shown in Figure \ref{fig:primitivedistribution}), but as the environments get more challenging, more passive primitives are introduced as well (Figure \ref{fig:primitivehist}).
What we observe from the histograms in Figure \ref{fig:primitivehist} is that new primitives are slowly introduced while the ranking of the primitives is also slightly changed.

\subsection{Active and Passive Primitive Frequencies}
\label{ap:primitive_frequencies}
\begin{figure*}[tb]
\centering
    \subfloat[Active Primitives]{
      \includegraphics[width=0.45\linewidth]{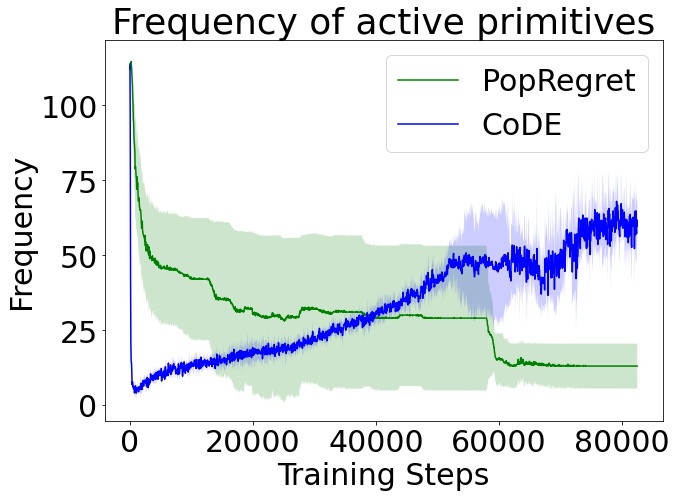}
    }%
    \subfloat[Passive Primitives]{\includegraphics[width=0.45\linewidth]{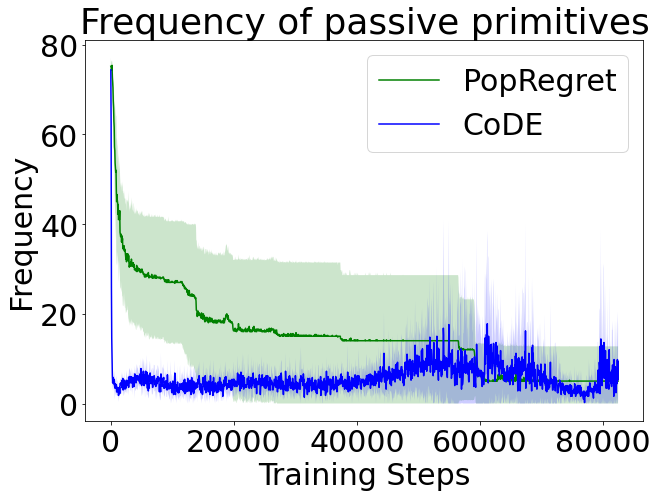}
    }
    \caption{Frequencies of active and passive primitives during training.}
    \label{fig:primitive_frequencies}
\end{figure*}

In Figure \ref{fig:primitive_frequencies}, we present frequencies of active and passive primitives during training.
With \WEDPb\, number of both active and passive primitives increase resulting in more complex websites.

\subsection{Creating primitives from DOM templates}
\label{ap:primitive-generation-algo}
\begin{algorithm}[tb]
  \textbf{Input:}{$D=(D_n, D_e)$: A DOM template, an HTML sub-tree with elements $D_n$ and edges $D_e$} 
  
  \textbf{Input:}{$V \subset D_n$: A list of elements that correspond to variables in $D_n$} 
  
 \textbf{Input:}{$A_{v,i}$: A list of variables $A_{v,i}$ for an element $v \in D_n$}
 
 \For {$v \in V$ \Comment{Iterate over elements.}}{
  Flip a coin. If it is heads, $D_n \longleftarrow D_n \setminus \{v\}$. \Comment{Add/remove an element.}
 }
 
 \For {$v \in D_n$ \Comment{Iterate over elements.}}{
 
 \For {$a \in A_{v,i}$ \Comment{Iterate over variables for element $v$.}}{
 Flip a coin. If it is heads, sample and assign a value for $a$. \Comment{Add/remove an variable.}
 }
 
  If there is at least one variable remaining for element $v$, $D_n \longleftarrow D_n \setminus \{v\}$.
 }
 \caption{Generating a new primitive from a DOM HTML template.}
 \label{alg:primitive-generation}
\end{algorithm}
In Algorithm \ref{alg:primitive-generation}, we outline the process for generating a new primitive from a given DOM HTML template.
A DOM template is a piece of self-contained HTML with variables.
We generate new primitives by assigning values to variables in DOM templates.

\subsection{Detailed Results on Test Environments}
\label{ap:detailed-results}
\begin{figure*}
    \small
    \begin{tabular}{c c c c c}
    & difficulty level = 1 & difficulty level = 2 & difficulty level = 3 & difficulty level = 4
    \\ 
    \raisebox{2.5\normalbaselineskip}[0pt][0pt]{\rotatebox[origin=c]{90}{\small Login}} & \includegraphics[width=0.22\linewidth]{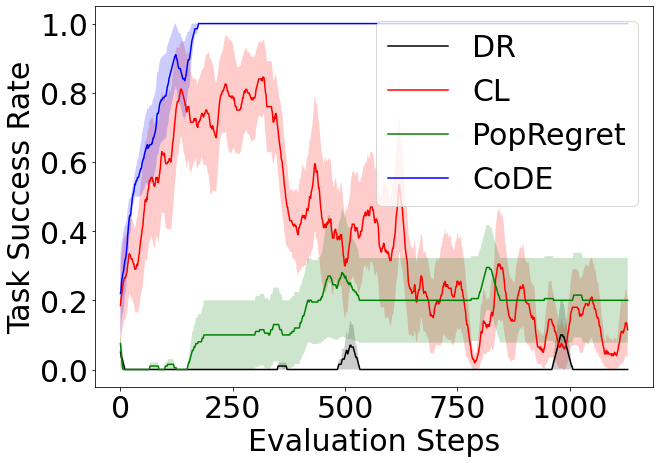}
    &\includegraphics[width=0.22\linewidth]{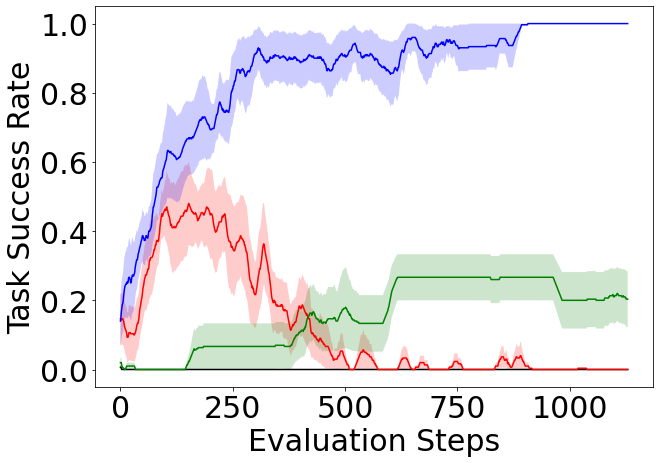}
    &\includegraphics[width=0.22\linewidth]{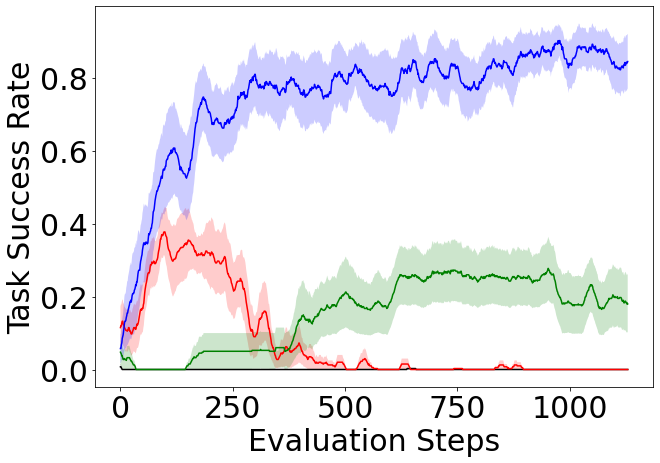}
    &\includegraphics[width=0.22\linewidth]{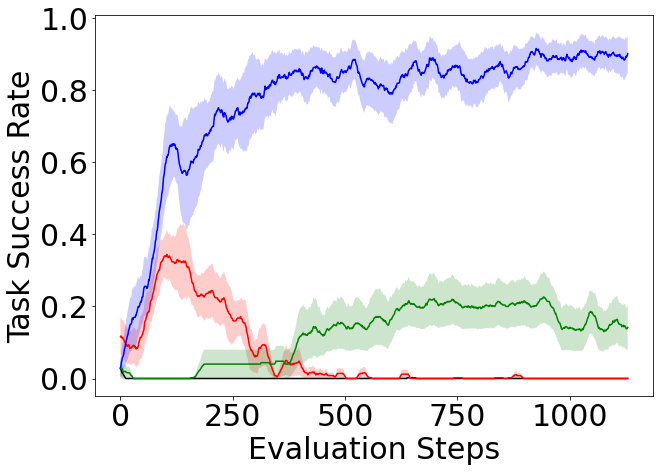}
    \\
    \raisebox{2.5\normalbaselineskip}[0pt][0pt]{\rotatebox[origin=c]{90}{\small Enter Address}} &\includegraphics[width=0.22\linewidth]{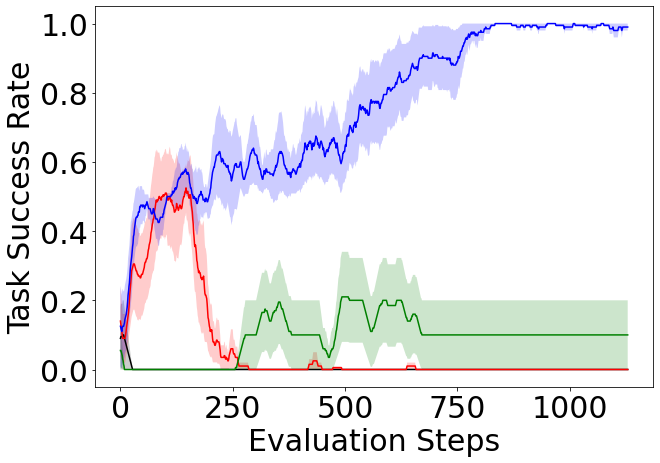}
    &\includegraphics[width=0.22\linewidth]{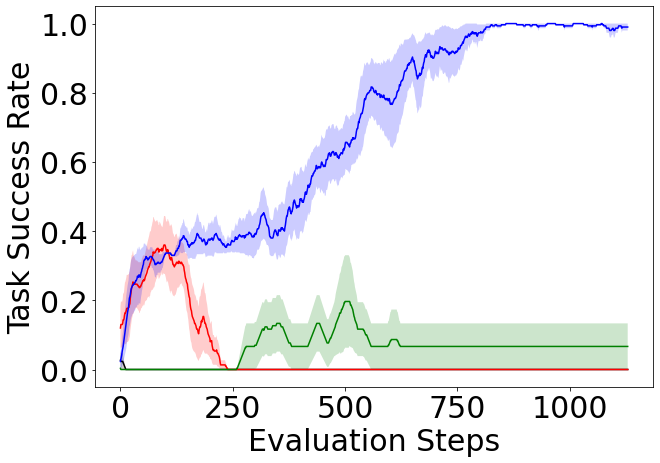}
    &\includegraphics[width=0.22\linewidth]{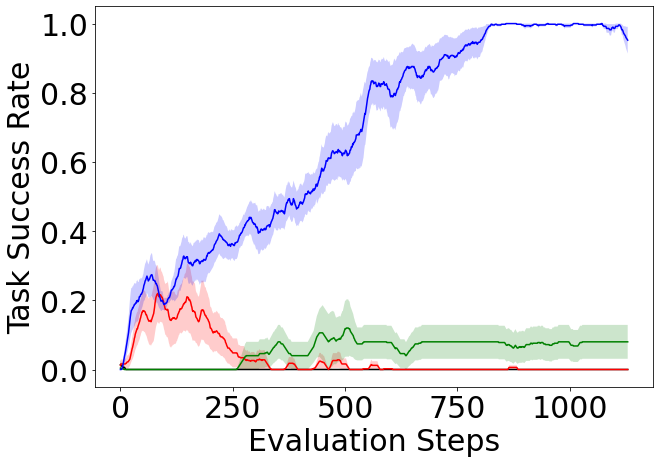}
    &\includegraphics[width=0.22\linewidth]{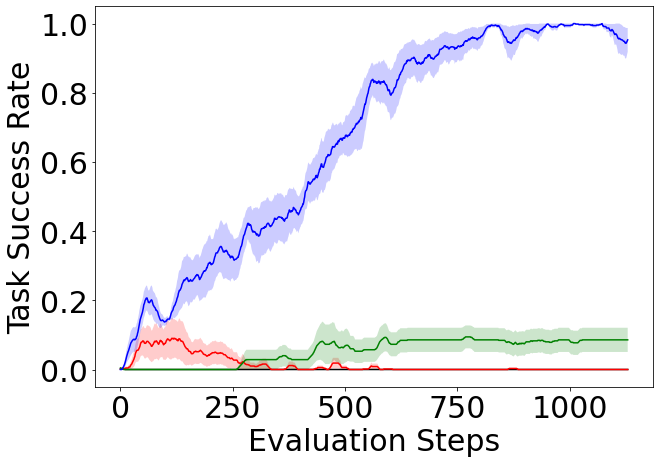}
    \\
    \raisebox{2.5\normalbaselineskip}[0pt][0pt]{\rotatebox[origin=c]{90}{\small Enter Payment}} &\includegraphics[width=0.22\linewidth]{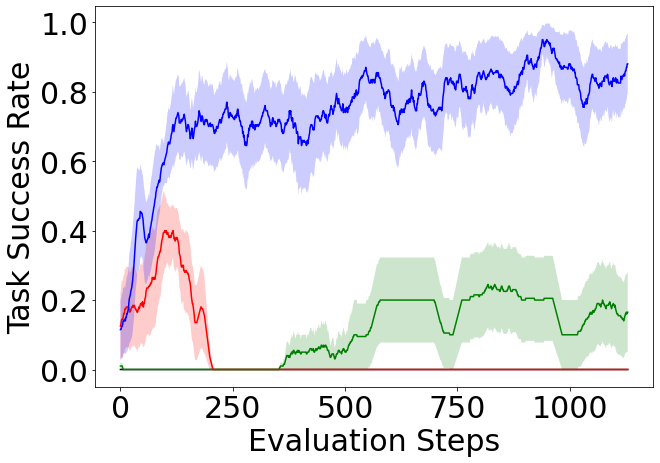}
    &\includegraphics[width=0.22\linewidth]{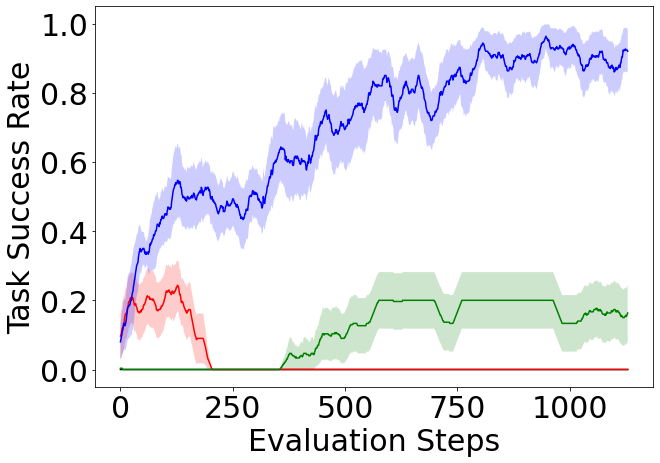}
    &\includegraphics[width=0.22\linewidth]{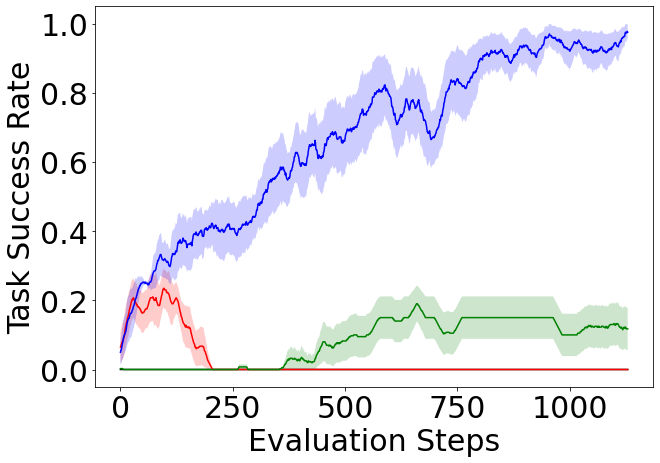}
    &\includegraphics[width=0.22\linewidth]{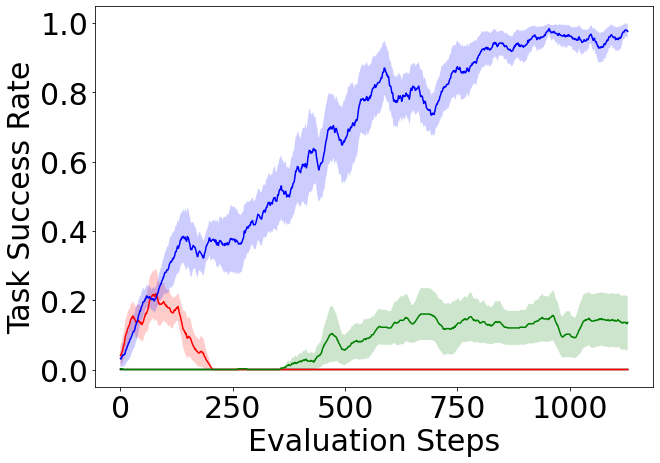}
    \\
    \raisebox{2.5\normalbaselineskip}[0pt][0pt]{\rotatebox[origin=c]{90}{\small Shopping}} &\includegraphics[width=0.22\linewidth]{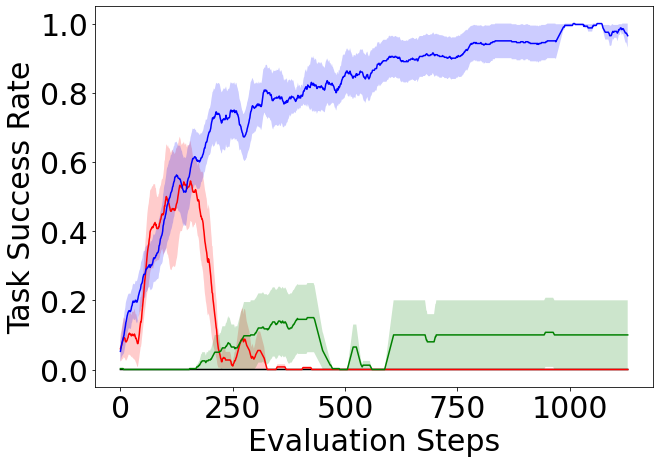}
    &\includegraphics[width=0.22\linewidth]{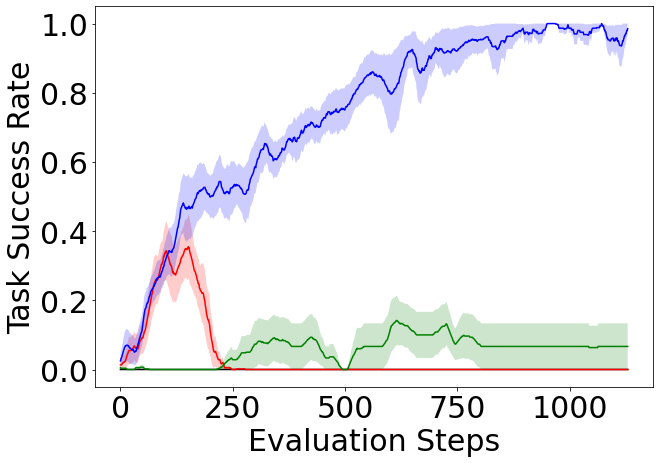}
    &\includegraphics[width=0.22\linewidth]{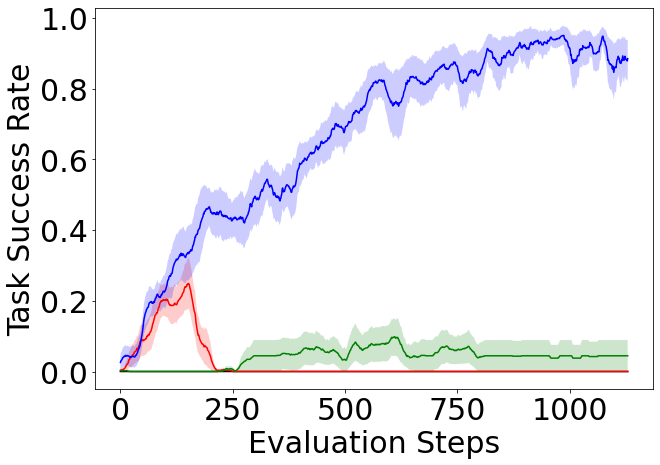}
    &\includegraphics[width=0.22\linewidth]{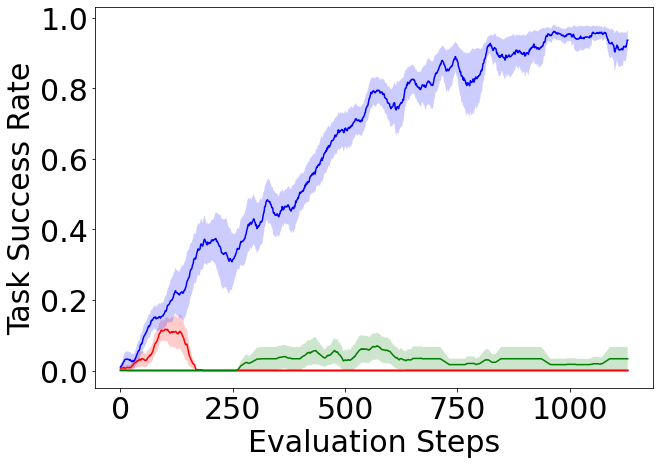}
    \\
    \raisebox{2.5\normalbaselineskip}[0pt][0pt]{\rotatebox[origin=c]{90}{\small Flight Booking}} &\includegraphics[width=0.22\linewidth]{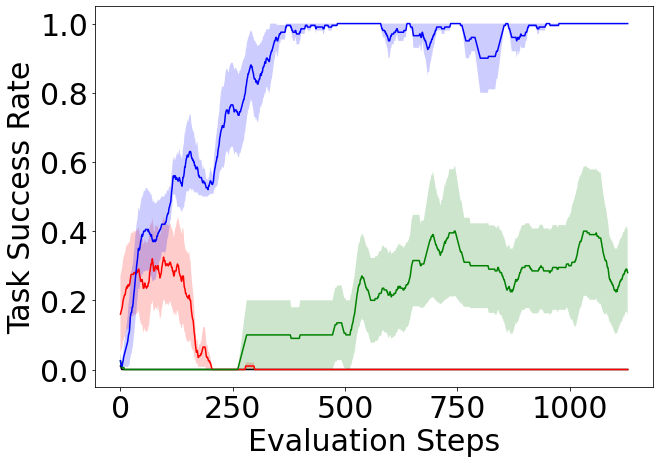}
    &\includegraphics[width=0.22\linewidth]{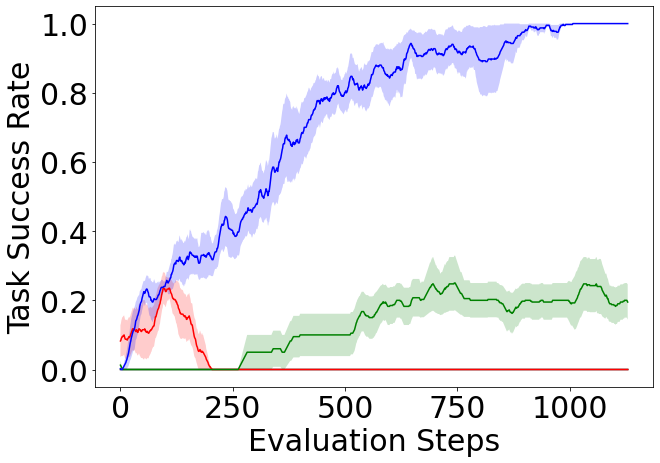}
    &\includegraphics[width=0.22\linewidth]{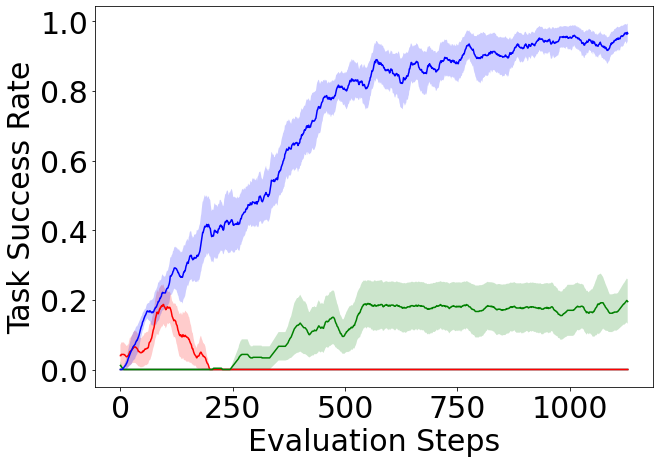}
    &\includegraphics[width=0.22\linewidth]{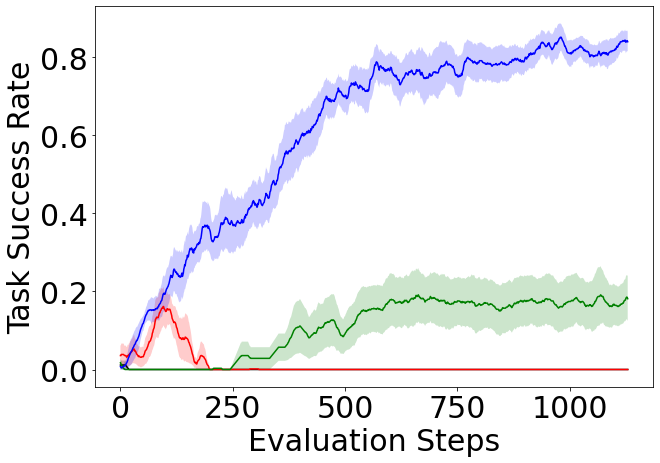}
    \\
    \end{tabular}
    \caption{Task success rate comparison of \WEDPb\ and baseline models on test environments with increasing difficulty levels. From left to right, columns correspond to increasing difficulty. From top to bottom, rows correspond to different test environments.}
    \label{fig:tasksuccess}
\end{figure*}

We detail the aggregated results in Figure \ref{fig:tasksuccessaggregated} and present performance of agents across tasks and difficulty levels (Figure \ref{fig:tasksuccess}).
On the easiest level of tasks, CL achieves slightly lower performance than \WEDPb\ early in the training while as the task difficulty increases, the gap becomes more apparent.

\subsection{cMiniGrid Details}
\begin{figure*}[tb]
    \subfloat[Global Subtask Workflow]{
      \includegraphics[width=0.6\linewidth]{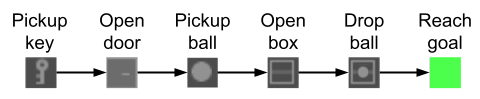}
    }%
    \subfloat[Sample Design Workflow]{
      \includegraphics[width=0.35\linewidth]{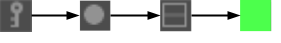}
    }
    \caption{The global subtask workflow (a) and a sample design (b). Sample design respects the dependency structure in the global subtask workflow.}
    \label{fig:minigrid-subtask-graphs}
\end{figure*}

In cMiniGrid, we use 5 different subtask primitives (Figure \ref{fig:minigrid-subtask-graphs}): (i) Pickup the key, (ii) Open the door, (iii) Pickup the ball, (iv) Open the box, and (v) Drop the ball.
The adversary designs a grid by selecting a set of subtasks and cMiniGrid stitches them together according to the global subtask workflow.
In Figure \ref{fig:minigrid-subtask-graphs}, we present the global subtask workflow and a sample design.
In this example, the adversary selects \textit{pickup key, pickup ball}, and \textit{open box} subtasks.
The sample workflow is generated from the set of subtasks while respecting the global workflow.
Finally, cMiniGrid randomly places corresponding objects and the agent to empty cells in the grid.
We assume that there is only a single room in the grid and the goal is always included in the set of subtasks.
We use the global subtask workflow as our testbed to evaluate the performance of the population of agents.

Different from designing web pages, the order of subtasks has no effect in cMiniGrid.
The adversary network is similar to the one used in gMiniWoB except that there are no predictions for number of cells and cell locations as objects are randomly assigned to cells.
To train the population of agents, we use a 3-layer Convolutional Neural Network (CNN) for image and a 2-layer Multi Layer Perceptron (MLP) for position observations, respectively.
A final 2-layer MLP is applied to image and position encodings to generate a distribution over actions.
We use ReLU activations in both networks.

\subsection{Comparison of $\alpha$ in budget weighting}
\label{ap:weight_comp}
\begin{figure*}[tb]
\small
    \subfloat{
      \includegraphics[width=0.8\linewidth]{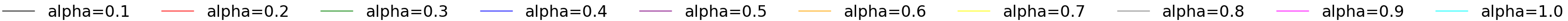}
    } \\
    \subfloat[Difficulty level 1]{
      \includegraphics[width=0.45\linewidth]{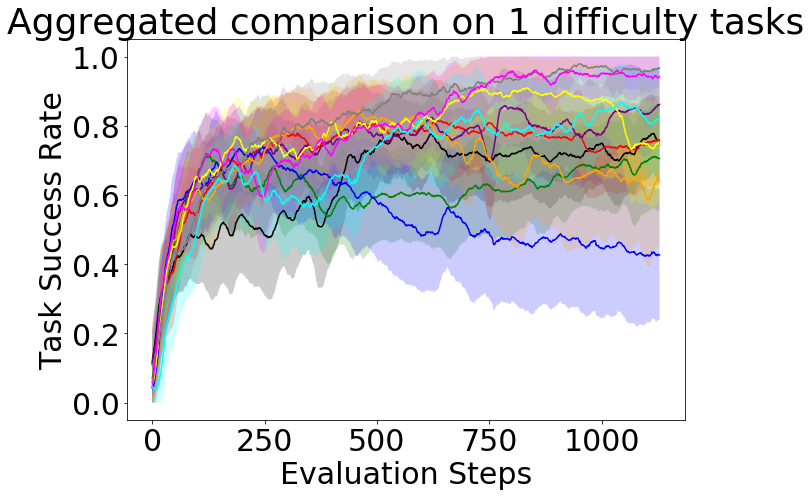}
    }%
    \subfloat[Difficulty level 2]{
      \includegraphics[width=0.45\linewidth]{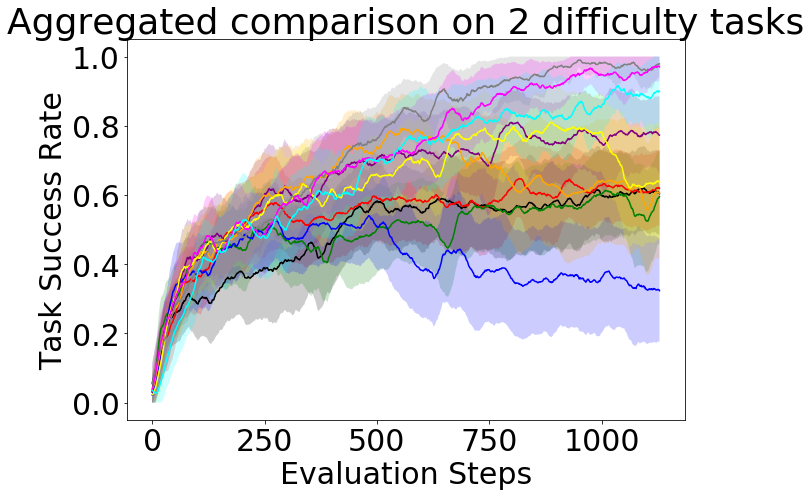}
    } \\
    \subfloat[Difficulty level 3]{
      \includegraphics[width=0.45\linewidth]{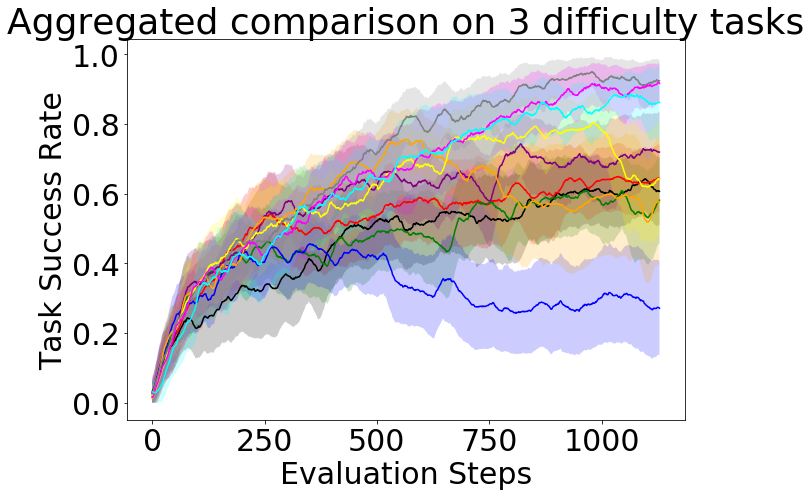}
    }%
    \subfloat[Difficulty level 4]{
      \includegraphics[width=0.45\linewidth]{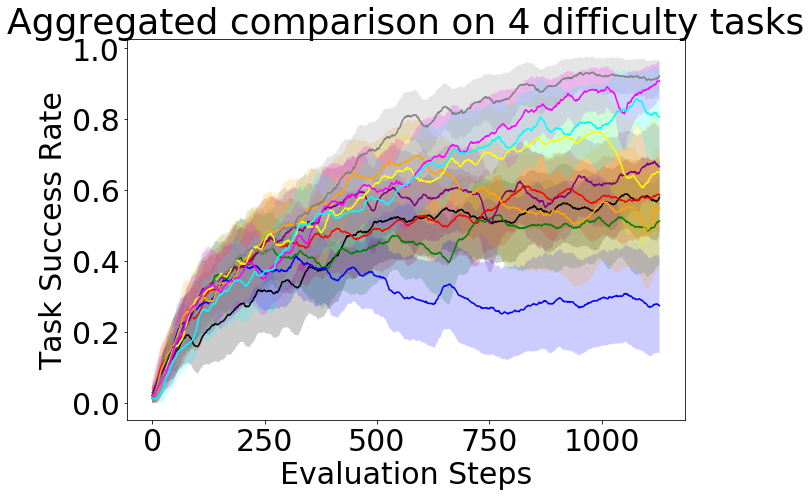}
    }%
    \caption{\small Aggregated task success rate comparison of \WEDPb\, trained with different $\alpha$ weights. \WEDPb\ yields the best performance with a strong $\alpha$ illustrating the importance of the introduced budget mechanism for the compositional task design in web navigation.}
    \label{fig:weight_comp}
    \vspace{-.7cm}
\end{figure*}

In Figure \ref{fig:weight_comp}, we plot results where \WEDPb\ is trained with different $\alpha$ weights.
We observe that as the $\alpha$ increases, we get consistent improvements illustrating the importance of the novel budget loss.
With very small $\alpha$, the performance also degrades over time as the adversary is not able to adapt to skill level of agents.

\subsection{Web Environment Design Primitives}
\label{ap:design_primitives}
\begin{table*}
\begin{center}
\begin{tabular}{ r|c|c|p{5cm}  }
 \multicolumn{4}{c}{\large \textbf{Design Primitives and Their Descriptions}} \\
 \hline 
 \textbf{Design Primitive} & \textbf{Design Template} & \textbf{Active/Passive} & \textbf{Description}\\
 \hline \hline
    addressline1   & input & active & Main address information\\ \hline
    addressline2 &  input  & active & Secondary address information\\ \hline
    cabin & multi-selection & active & Multiple cabin options\\ \hline
    captcha & input & active & Captcha information\\ \hline
    carousel & carousel & passive & Items with images in a carousel with previous and next buttons \\ \hline
    cart & cart & passive & Items in a product cart with promo code information\\ \hline
    cc & multi-selection & active & Multiple credit card type options\\ \hline
    cccvv & input & active & Credit card CVV information\\ \hline
    ccexpdate & input & active & Credit card expiration date information \\ \hline
    ccnumber & input & active & Credit card number information\\ \hline
    city & input & active & City address information\\ \hline
    dealmedia & media & passive & Product media with image, label, and link\\ \hline
    deck & deck & passive & Multiple product decks with image, label, and link\\ \hline
    departureairport & input & active & Departure airport information \\ \hline
    departuredate & input & active & Departure date information\\ \hline
    destinationairport & input & active & Destination airport information \\ \hline
    destinationdate & input & active & Destination date information\\ \hline
    firstname & input & active & First name information\\ \hline
    flighttype & multi-selection & active & Multiple flight type options\\ \hline
    footer & footer & passive & Footer with links and information\\ \hline
    forgotpassword & link & passive & Link with forgot password context \\ \hline
    forgotusername & link & passive & Link with forgot username context\\ \hline
    fullname & input & active & First and last name information\\ \hline
    header & label & passive & Generic header\\  \hline
    header\_login & label & passive & Header for login form\\ \hline
    header\_select\_items & label & passive & Header for item selection\\ \hline
    inpgroup & input & passive & Generic input with default search context\\ \hline
    lastname & input & active & Last name information\\ \hline
    navbar & navigation bar & passive & A navigation bar with a menu \\ \hline
    next\_checkout & button & passive & Next button with checkout context \\ \hline
    next\_login & button & passive & Next button with login context \\ \hline
    next\_login\_page & button & passive & Next button with login context \\ \hline
    numberofpeople & multi-selection & active & Multiple number of people options\\ \hline
    password & input & active & Password information \\ \hline
    rememberme & selection & active & Checkbox with remember me context\\ \hline
    state & input & active & State information \\ \hline
    stayloggedin & selection & active & Checkbox with stay logged in context\\ \hline
    submit & button & passive & Submit button \\ \hline
    username & input & active & Username information\\ \hline
    zipcode & input & active & Zipcode information \\ \hline
\end{tabular}
\caption{}\label{table:primitives}
\end{center}
\end{table*}
In Table \ref{table:primitives}, we present the list of design primitives, corresponding templates, types, and descriptions.

\subsection{List of Test Environments}
\label{ap:test}
\begin{figure*}
    \centering
    \subfloat[Home Page]{
      \includegraphics[width=0.23\linewidth]{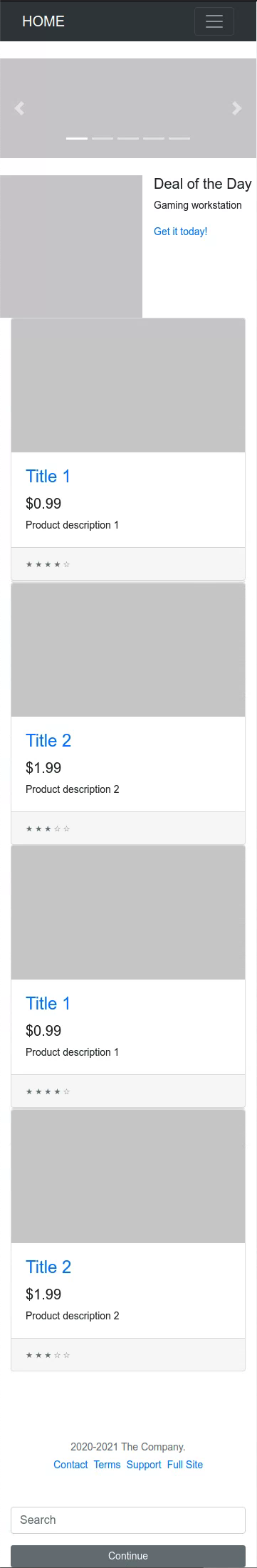}
    }%
    \subfloat[Login Page]{
      \includegraphics[width=0.23\linewidth]{img/sampledesigns/logintask.png}
    }%
    \subfloat[Address Page]{
      \includegraphics[width=0.23\linewidth]{img/sampledesigns/addresstask.png}
    }%
    \caption{Screenshots of multi-page ``Shopping'' environment. The ``Shopping'' environment is composed of a complex home page and additional ``Login'' and ``Enter Address'' pages.}
    \label{fig:shoppingtestenv}
\end{figure*}

In Figure \ref{fig:shoppingtestenv}, we present screenshots of the testing environments with the hardest difficulty levels.
While ``Login'', ``Enter Address'', ``Enter Payment'', and ``Flight Booking'' are single page environments, ``Shopping'' is a multi-page environment where an agent needs to first navigate the home page and then solve ``Login'' and ``Enter Address'' tasks.
\begin{figure*}
    \subfloat[Login]{
      \includegraphics[width=0.23\linewidth]{img/sampledesigns/logintask.png}
    }%
    \subfloat[Enter Address]{
      \includegraphics[width=0.23\linewidth]{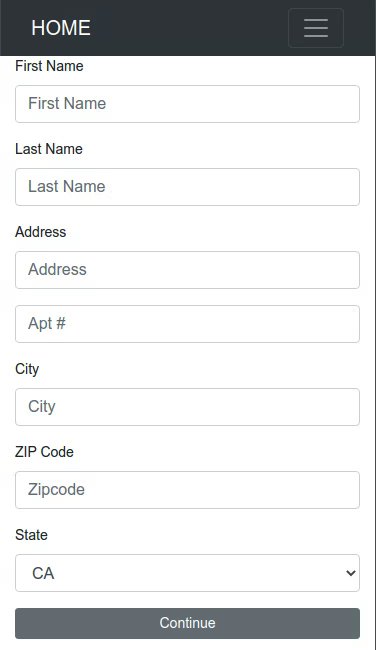}
    }%
    \subfloat[Enter Payment]{
      \includegraphics[width=0.23\linewidth]{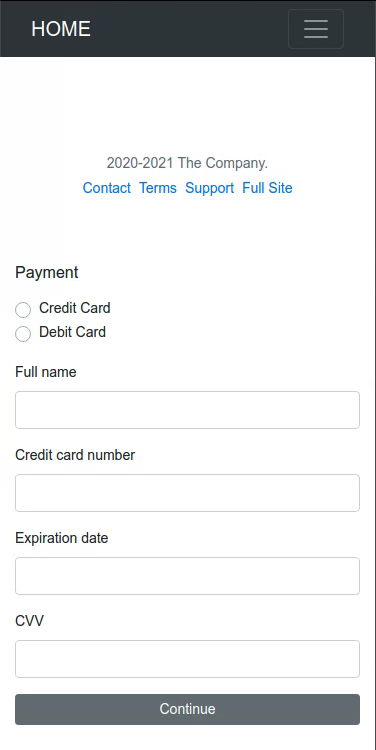}
    }%
    \subfloat[Flight Booking]{
      \includegraphics[width=0.23\linewidth]{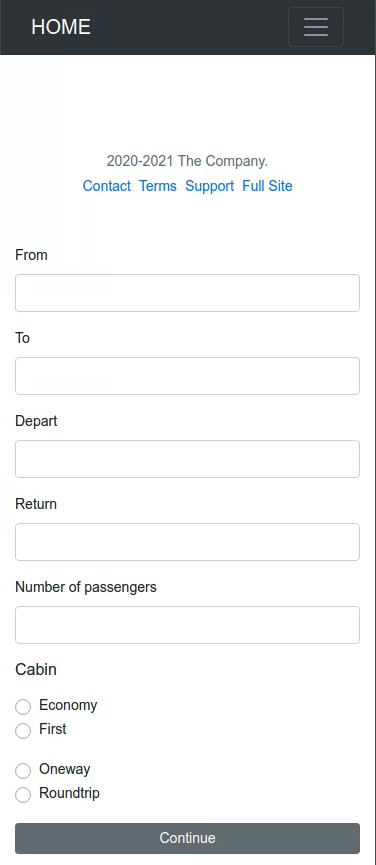}
    }%
    \caption{Screenshots of single page test environments.}
\end{figure*}

\subsection{Example web page designs}
\label{ap:example_designs}
\begin{figure*}
    \subfloat{
      \includegraphics[width=0.23\linewidth]{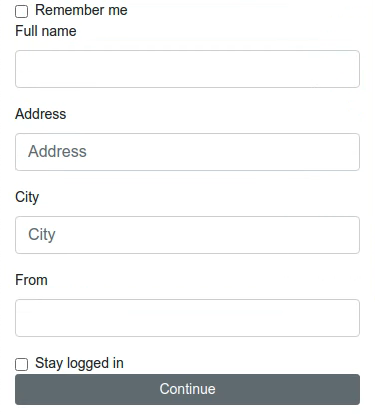}
    }%
    \subfloat{
      \includegraphics[width=0.23\linewidth]{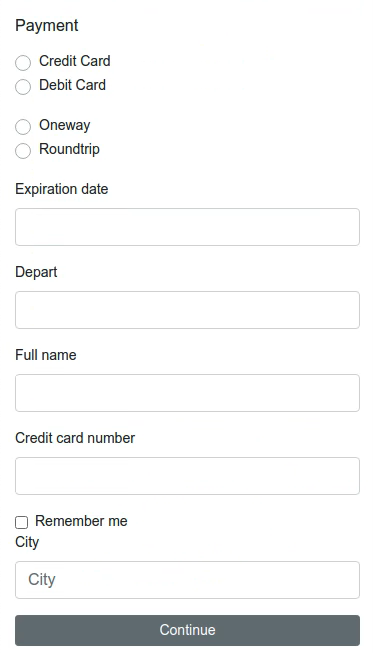}
    }%
    \subfloat{
      \includegraphics[width=0.23\linewidth]{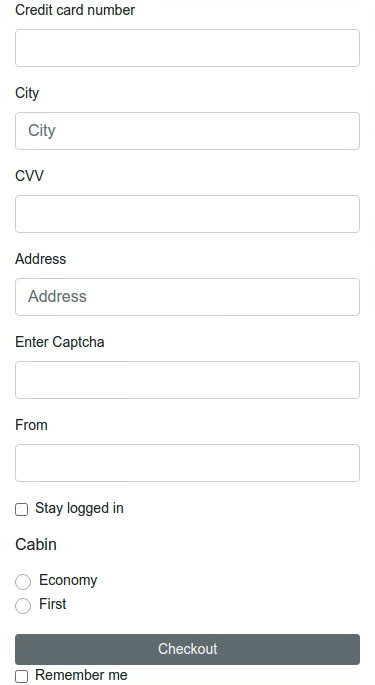}
    }%
    \subfloat{
      \includegraphics[width=0.23\linewidth]{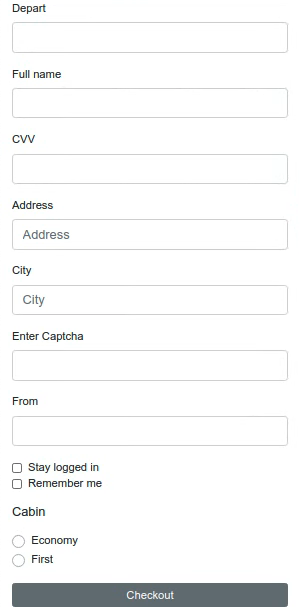}
    }\\
    \subfloat{
      \includegraphics[width=0.23\linewidth]{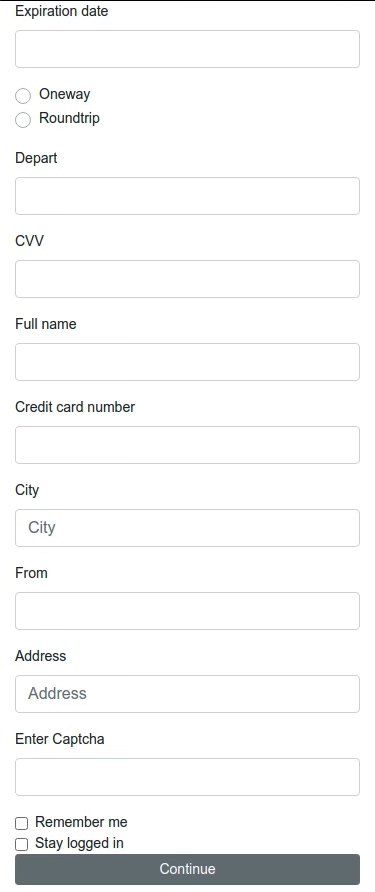}
    }%
     \subfloat{
      \includegraphics[width=0.23\linewidth]{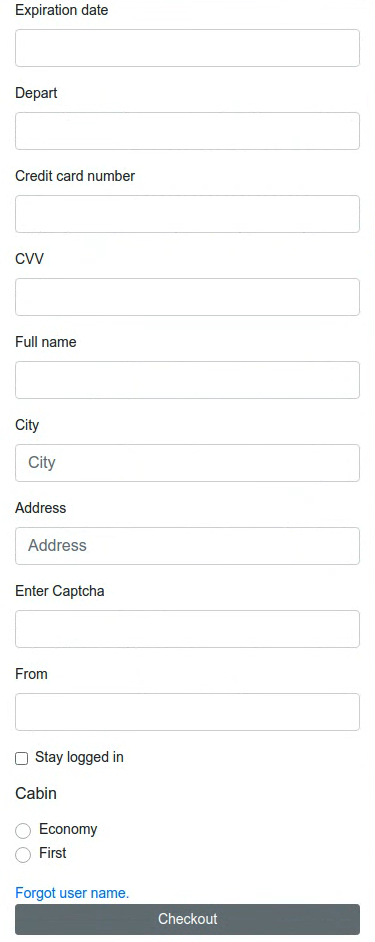}
    }%
    \subfloat{
      \includegraphics[width=0.23\linewidth]{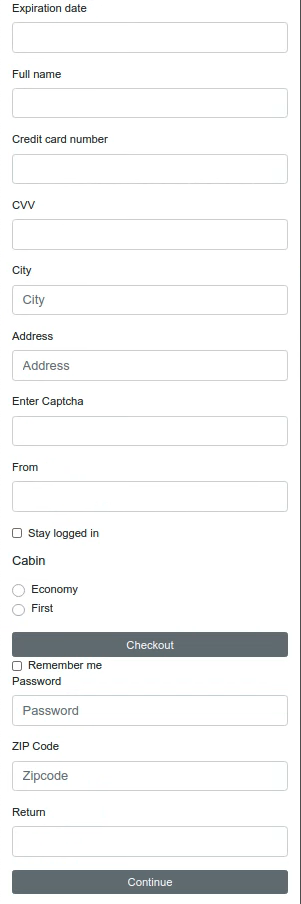}
    }%
    \subfloat{
      \includegraphics[width=0.23\linewidth]{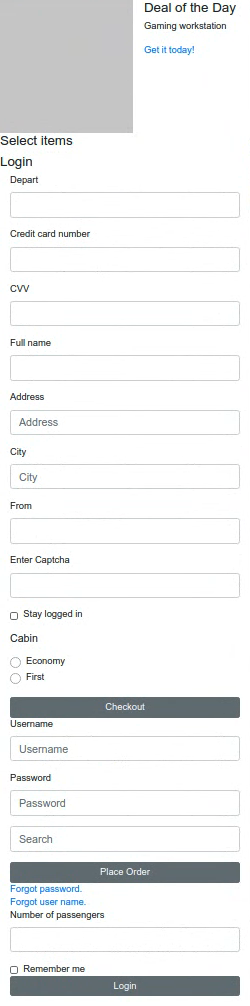}
    }%
    \caption{Screenshots of sample pages generated by the adversary.}
    \label{fig:testenvs}
\end{figure*}

In Figure \ref{fig:testenvs}, we present more screenshots of generated pages by the adversary from including multi-page websites.
They cover a very broad spectrum of complexities and DOM tree structures.
As an example, two web pages on the top right both have "City", "CVV", and "Address" elements but with different orders.
This allows the web navigation agents to observe a website in multiple different ways for better generalization.

\subsection{Implementation Details on Web navigation and adversary networks}
\label{ap:adversary_network_details}
Following \cite{gur2018learning}, we design web navigation agent networks as DOM and profile encoders with pairwise similarity scoring.
Each web navigation agent policy network has 104501 parameters.

\begin{figure*}
    \centering
    \includegraphics[scale=0.6]{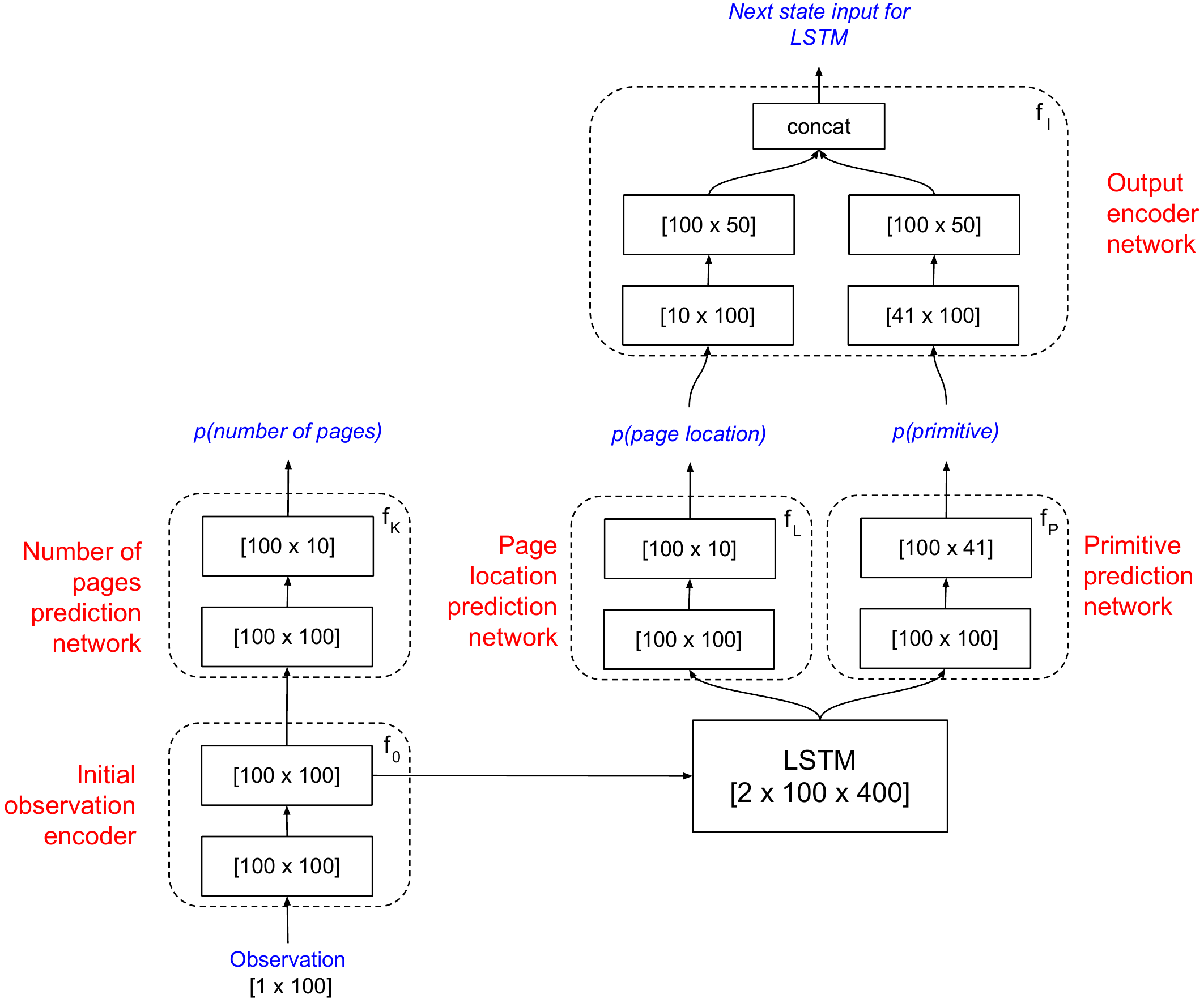}
    \caption{Adversary network architecture. Each box corresponds to a dense layer with the shape of the corresponding linear transformation matrix. Each dense layer also includes a bias vector with the same size as the columns of the corresponding matrices.}
    \label{fig:adversary_network}
    \vspace{-0.4cm}
\end{figure*}

In Figure \ref{fig:adversary_network}, we detail the adversary network architecture for a single design action with the parameters used in this work.
We use 100 dimensions for hidden vectors for all dense layers as well as the LSTM network.
Every dense layer is stacked twice and tanh activation function is applied on the output of all non-final dense layers.
Total number of parameters for the adversary policy network is 152461.

For both cMiniGrid and gMiniWoB tasks, we use $\beta=\delta=0.0$ for reward thresholds and $M=2$ based on a hyper-parameter search (see Appendix \ref{ap:ablation}).

\subsection{Limitations}
\label{app:limitations}
One limitation of our method is that the difficulty objective in Eq. \ref{eq:budget} is based on the number of primitives $N$ added to the compositional task. It may not always be the case that difficulty is a direct function of the number of primitives, and in that case the difficulty objective would not generalize to those types of tasks. As an alternative incentive, we provide the population-based regret (PopRegret). Although the empirical results show that optimizing for PopRegret alone learns much more slowly, it is more environment-agnostic and could be useful for a range of tasks.

A second limitation of our method is that it still requires building an environment-specific rendering function that can translate the \adversary's actions into a feasible compositional task (e.g. by linking web pages together). We rely on pre-defined primitives for each environment to enable the adversary to construct the tasks. 

\subsection{Broader Impact}
\label{app:broader_impact}
The immediate application of this work is to train RL agents to complete web navigation tasks for users. This could free up time for people, because instead of manually clicking through websites to perform tasks such as flight booking, they could simply issue a language query such as ``Book me a flight to San Francisco on Tuesday". We do not foresee that this application would lead to job automation, since by its nature web-based form-filling is a task that does not require interacting with a person. More broadly, enabling RL agents to better perform compositional tasks could be a step towards future applications such as autonomous driving and household robotics. Each of these applications comes with both potential harms (such as job automation) and benefits (such as increased efficiency, or improved elder care).

\end{document}

%% file: math_commands.tex
\usepackage{amsmath,amsfonts,bm}
\DeclareMathOperator*{\E}{\mathbb{E}}

\def\eqref#1{equation~\ref{#1}}

\def\1{\bm{1}}

\DeclareMathAlphabet{\mathsfit}{\encodingdefault}{\sfdefault}{m}{sl}
\SetMathAlphabet{\mathsfit}{bold}{\encodingdefault}{\sfdefault}{bx}{n}

\renewcommand{\E}{\mathbb{E}}

\DeclareMathOperator*{\argmax}{arg\,max}